%% file: main.tex
\documentclass[11pt,a4paper]{article}

\usepackage{xpeng-template}

\usepackage{hyperref}
\usepackage{float}

\input{main/macros}

\xpengcorrespondence{Jie Chen}{chenj81@xiaopeng.com}

\title{%
{\LARGE Athena-Brain Technical Report}\\[0.4em]
{\large An Efficient Robot Brain for General Intelligence and Embodied Interaction}
}

\author{%
  Jialian Li, Junhong Liu, Yuchen Cao, Weiran Guo, Jiaming Song, Xutao Wang, Yi Zhao, Jiangpin Liu, Jie Chen$^{\dagger}$%
}
\date{}  

\begin{document}

\begin{xpenghero}
  \input{main/abstract}
\end{xpenghero}

\begin{figure}[H]
    \centering \includegraphics[width=.965\linewidth]{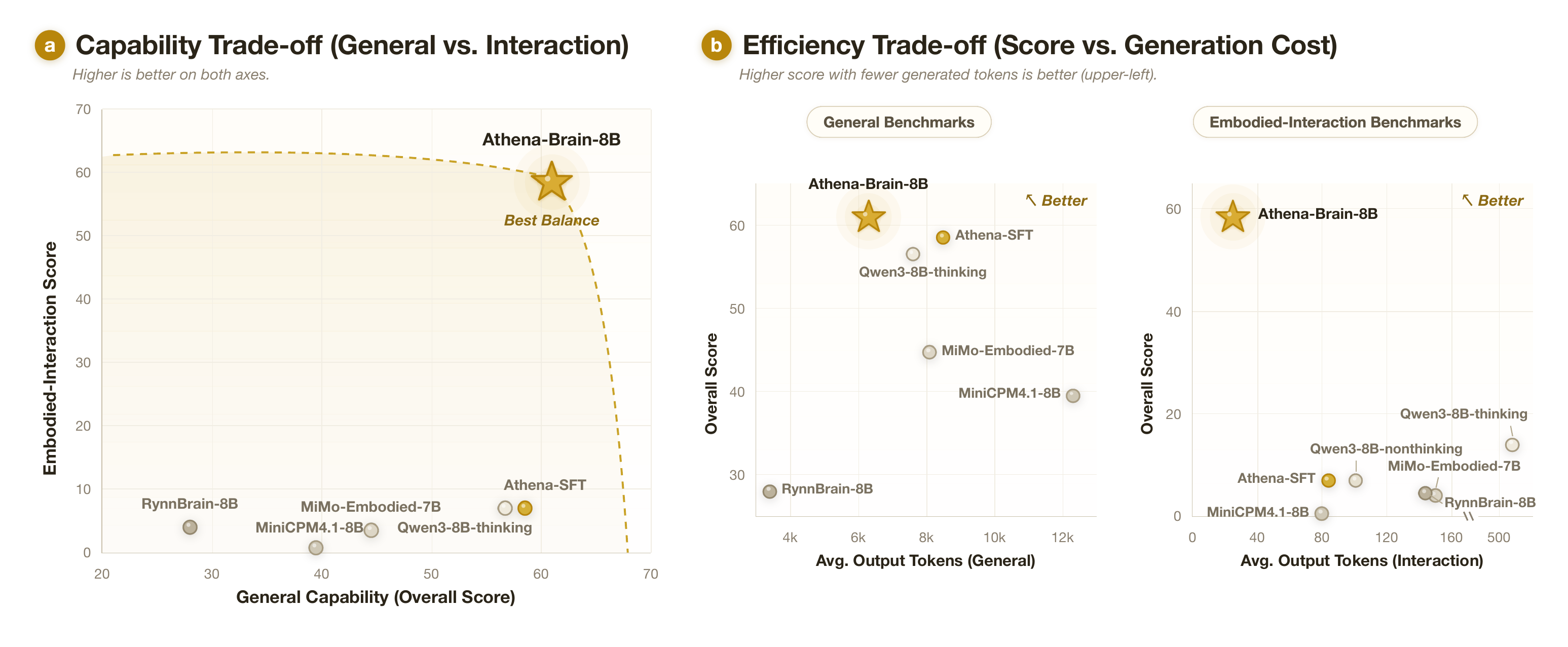}
    \caption{Overall comparison of Athena-Brain-8B with representative open-source models across general capabilities, embodied capabilities, and generation efficiency. Athena-Brain-8B is designed to balance these three objectives within a compact 8B language model. Athena-Brain-8B maintains competitive general language performance, achieves strong embodied capabilities among models of similar scale, and generates substantially shorter responses than its corresponding thinking counterpart, demonstrating an effective balance between capability and generation efficiency.}
    \label{fig:pareto}
\end{figure}

\input{main/introduction}
\input{main/athena_at_a_glance}
\input{main/training_pipeline}
\input{main/evaluation}

\input{main/analysis}
\input{main/related_work}
\input{main/conclusion}

\bibliographystyle{unsrtnat}
\bibliography{refs}



\end{document}

%% file: main/macros.tex
\usepackage{placeins}



%% file: main/abstract.tex
\begin{abstract}

Large language models (LLMs) have demonstrated remarkable capabilities in language understanding, reasoning, and world knowledge. As embodied agents become increasingly capable, there is a growing demand for compact models that can serve as an on-device brain, preserving the broad general intelligence of LLMs while enabling effective high-level interaction with embodied environments. Existing approaches, however, often prioritize either general-purpose intelligence or specialized embodied capabilities, making it challenging to satisfy both requirements within a single model.
We present \textbf{Athena-Brain-8B}, an 8B LLM designed to serve as an on-device brain for embodied intelligence for embodied intelligence. Through a multi-stage post-training pipeline consisting of General Supervised Fine-Tuning, General Reinforcement Learning, Embodied Expert training, and Model Merge, Athena-Brain-8B maintains strong general capabilities while acquiring strong high-level embodied interaction capabilities and generating concise responses for efficient embodied interaction.
Experimental results demonstrate the effectiveness of Athena across both general and embodied evaluations. Compared with the corresponding Qwen3-8B thinking model~\citep{yang2025qwen3technicalreport}, Athena-Brain-8B achieves comparable performance on general language and reasoning benchmarks while generating substantially shorter responses. On in-domain embodied benchmarks, Athena-Brain-8B consistently outperforms models of similar scale and surpasses several substantially larger frontier models evaluated zero-shot, demonstrating that compact language models can effectively integrate strong general intelligence with embodied capabilities.
\end{abstract}

%% file: main/introduction.tex
\section{Introduction}

Large language models (LLMs) have demonstrated remarkable capabilities in language understanding, reasoning, and world knowledge~\citep{openai2026gpt55,anthropic2026claudesonnet5,qwen2026qwen37max,zhipuai2026glm52}. As embodied agents become increasingly capable and practical, there is a growing demand for compact language models that can serve as an on-device brain, supporting high-level understanding, reasoning, planning, decision making, and interaction in embodied environments\citep{zitkovich2023rt,driess2023palm}. Such a robot brain should not only preserve the broad general capabilities of foundation LLMs, but also acquire robust embodied capabilities for interacting with embodied environments, while maintaining efficient responses to support responsive and natural interaction.

Recent embodied foundation models have substantially advanced embodied intelligence by specializing in domain-specific interaction tasks~\cite{hao2025mimoembodiedxembodiedfoundationmodel,dang2026rynnbrain}. While effective for embodied benchmarks, these models are primarily optimized for learning embodied behaviors within specific environments, with less emphasis on the high-level reasoning and planning capabilities required of a general robot brain. In contrast, robots deployed in open-world settings must continually interpret instructions, decompose long-horizon tasks, actively explore partially observable environments, and interact with users or external tools. These capabilities rely heavily on the general intelligence developed by modern LLMs.

Meanwhile, frontier LLMs have achieved remarkable progress in language reasoning, dialogue, and computer-use agents~\cite{anthropic2024computeruse,openai2025chatgptagent,qin2025uitars}. However, they are not explicitly optimized for embodied interaction. Their reasoning-oriented inference often generates unnecessarily verbose responses, increasing interaction latency, and they lack post-training tailored for embodied decision making. These limitations motivate the development of a compact robot brain that unifies broad general intelligence, high-level embodied interaction, and efficient generation within a single post-training framework.

To address these challenges, we present \textbf{Athena-Brain-8B}, an 8B large language model designed with on-device deployment as a target for embodied intelligence. Athena-Brain-8B maintains strong general capabilities while acquiring robust embodied capabilities, and is further optimized to generate concise responses for efficient embodied interaction. These objectives are achieved through a multi-stage post-training pipeline consisting of General Supervised Fine-Tuning, General Reinforcement Learning, Embodied Expert Training, and Model Merge.

Extensive experiments validate the effectiveness of Athena-Brain-8B across both general and embodied evaluations. Compared with the corresponding Qwen3-8B thinking model, Athena-Brain-8B achieves comparable or better performance on general language and reasoning benchmarks while requiring substantially fewer generated tokens. Since practical inference latency depends heavily on deployment hardware and system implementation, we use response generation length as a hardware-independent measure of generation efficiency throughout this report. On in-domain embodied benchmarks, Athena-Brain-8B consistently outperforms existing compact embodied models and even approaches or surpasses substantially larger models, demonstrating that a compact language model can effectively preserve strong general capabilities while acquiring robust embodied capabilities through efficient generation.

The remainder of this report is organized as follows. Section~2 provides an overview of the performance of Athena-Brain-8B. Section~3 describes the complete post-training pipeline that enables Athena-Brain-8B to balance general capabilities, embodied capabilities, and interaction efficiency. Section~4 presents comprehensive evaluations on both general language and embodied benchmarks. Section 5 provides detailed analyses of the proposed training strategy, offering insights into the effectiveness of different training stages and design choices. Finally, Section~6 reviews the related work and Section 7 concludes the report and discusses future research directions.

%% file: main/athena_at_a_glance.tex
\section{Athena-Brain-8B at a Glance}

Athena-Brain-8B is an 8B large language model designed to serve as an on-device brain for embodied intelligence. Built upon a multi-stage post-training pipeline, Athena-Brain-8B is designed to preserve strong general-purpose capabilities while acquiring robust embodied capabilities and enabling concise response generation for efficient embodied interaction.

We compare Athena-Brain-8B with three categories of representative open-source compact language models. Qwen3-8B serves as a strong general-purpose baseline built upon the same base model. The intermediate supervised fine-tuned model, Athena-SFT, is included to highlight the contribution of the subsequent post-training stages. We further compare against recent 7B--8B embodied and edge-oriented language models designed for robotic applications~\citep{minicpm4,hao2025mimoembodiedxembodiedfoundationmodel,dang2026rynnbrain}.
Figure~\ref{fig:pareto} provides an overall comparison of Athena-Brain-8B with representative open-source models across general capabilities, embodied capabilities, and generation efficiency. Athena-Brain-8B achieves the highest overall score on the selected general benchmarks while reducing the average response length compared with both the underlying Qwen3-8B and the intermediate SFT model. On in-domain embodied evaluations, Athena-Brain-8B substantially outperforms existing compact embodied models while requiring significantly fewer generated tokens, demonstrating an effective balance between general capabilities, embodied capabilities, and generation efficiency.

\begin{figure}[t]
    \centering \includegraphics[width=\linewidth]{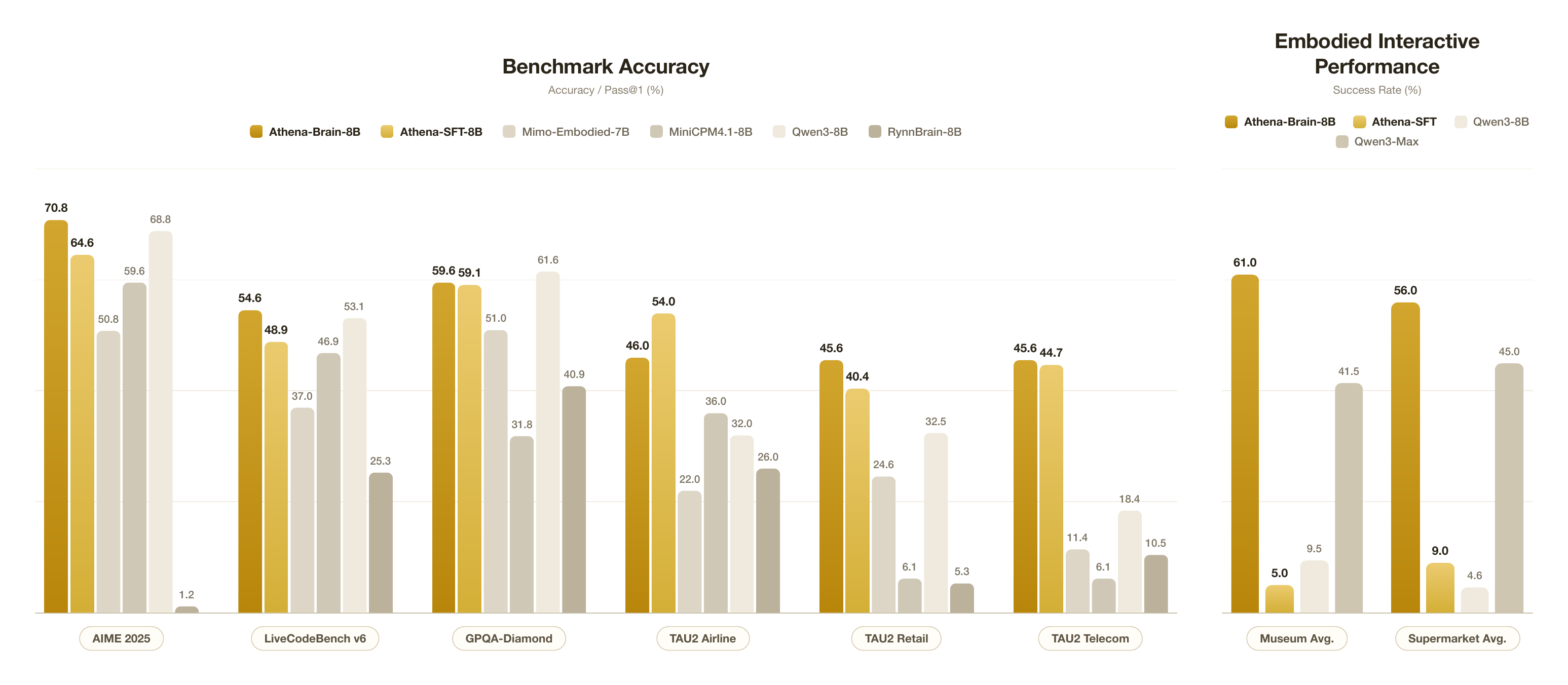}
    \caption{Representative performance of Athena-Brain-8B on selected general language and embodied interaction benchmarks. Athena-Brain-8B maintains competitive general language capabilities while substantially outperforming existing compact embodied language models on in-domain embodied interaction tasks. Notably, Athena-Brain-8B also achieves competitive or superior performance to several frontier large language models on in-domain embodied benchmarks, demonstrating the effectiveness of our post-training pipeline.}
    \label{fig:main_results}
\end{figure}

To provide a more intuitive view of Athena-Brain-8B's capabilities, Figure~\ref{fig:main_results} reports representative results on selected general and embodied benchmarks. Athena-Brain-8B consistently achieves competitive performance on general language understanding and reasoning tasks, showing that embodied post-training preserves the broad capabilities of the base LLM. More importantly, Athena-Brain-8B significantly outperforms existing open-source models of similar scale on our indomain embodied interaction benchmarks, and even surpasses substantially larger models such as Qwen3-Max~\citep{qwen2025qwen3max}. These results demonstrate that Athena-Brain-8B effectively combines strong general intelligence with strong in-domain text-based embodied interaction performance within a compact 8B language model. Detailed evaluation results can be found in Table~\ref{tab:sft-full-results} and \ref{tab:interactive_indomain}. 

%% file: main/training_pipeline.tex
\section{Training Pipeline}

Athena-Brain-8B is developed through a multi-stage post-training pipeline that progressively builds the capabilities required for a practical robot brain. Rather than optimizing all objectives simultaneously, we decompose the training process into four sequential stages, where each stage focuses on developing a specific aspect of the model while naturally serving as the foundation for the subsequent stage. This progressive strategy enables Athena to preserve strong general language intelligence, acquire robust embodied expertise, and ultimately integrate both capabilities into a unified model.

\begin{figure}[t]
    \centering
    \includegraphics[width=1\linewidth]{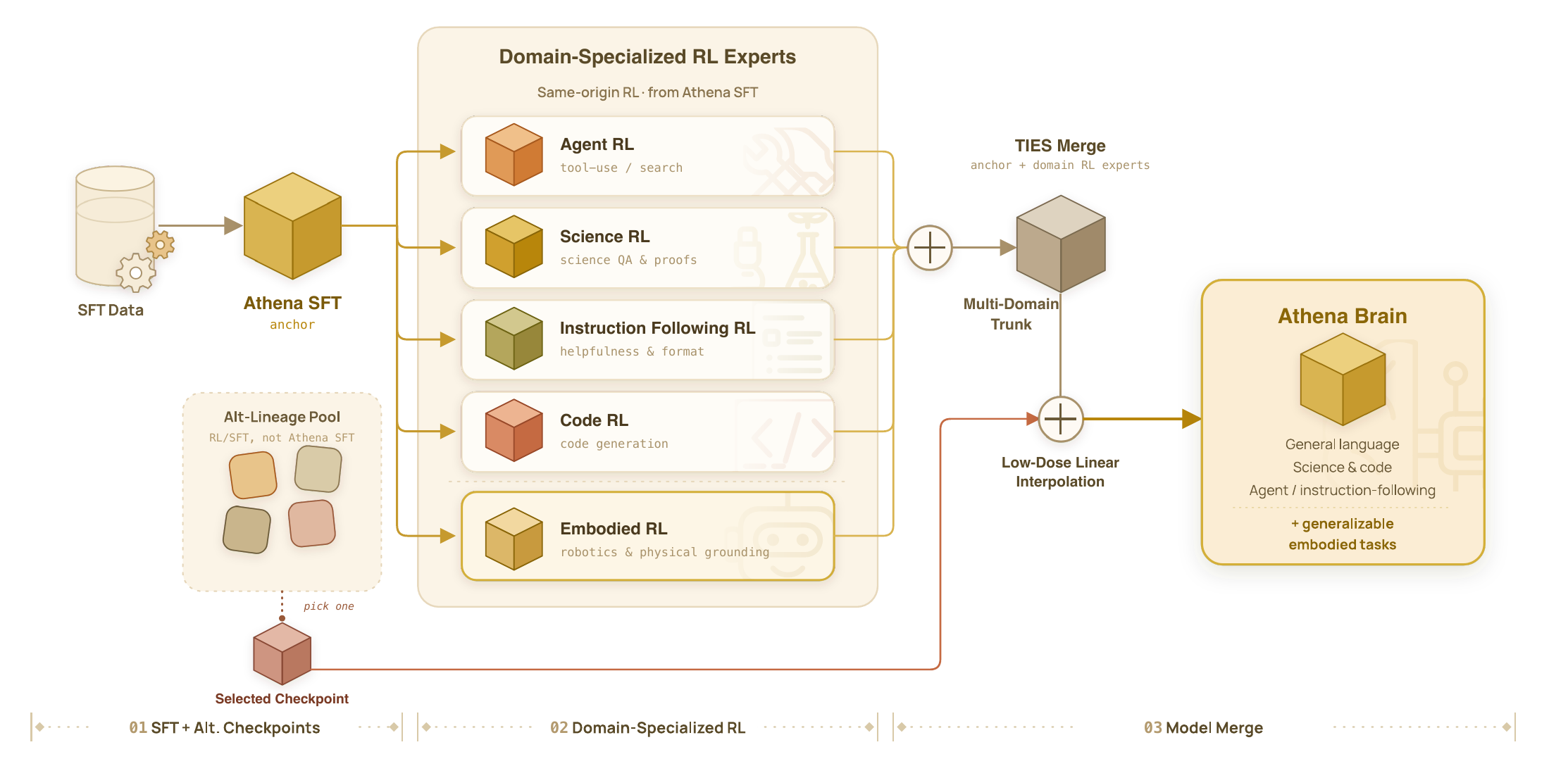}
    \caption{
    Overview of the Athena-Brain-8B post-training pipeline. Starting from an open-weight base model, we build a general-capability foundation, train capability-specialized and embodied-interaction experts, and merge them into a single compact robot-brain model for on-device deployment.
    }
    \label{fig:pipeline}
\end{figure}

As illustrated in Figure~\ref{fig:pipeline}, the training pipeline consists of four stages. We first perform \textbf{General Supervised Fine-Tuning (General SFT)} to establish high-quality instruction-following and reasoning capabilities. Building upon this foundation, \textbf{General Reinforcement Learning (General RL)} further improves reasoning performance and response quality through large-scale reinforcement learning. We then introduce an \textbf{Embodied Expert} stage, where the model is trained to develop robot-specific perception, planning, and decision-making abilities through embodied supervision and interactive environments. Finally, we perform \textbf{Model Merge} to integrate the general and embodied experts into a single unified model, achieving a favorable balance between general intelligence and embodied capability.

\subsection{General Supervised Fine-Tuning}

The first stage of Athena's training pipeline aims to establish a strong general-purpose foundation for subsequent capability development. Starting from the pretrained Qwen3-8B-Base model~\citep{yang2025qwen3technicalreport}, we perform supervised fine-tuning to improve instruction following, reasoning, coding, and general problem-solving capabilities while preserving the broad knowledge acquired during pretraining. Rather than relying on a single-stage training process, we adopt a progressive two-stage supervised fine-tuning strategy that balances broad capability acquisition with response quality refinement.

In the first stage, we perform large-scale supervised fine-tuning on a diverse instruction corpus covering a wide range of general capabilities. As summarized in Table~\ref{tab:sft-stage1-mixture}, this stage emphasizes comprehensive capability coverage, enabling the model to acquire robust instruction-following, reasoning, coding, multilingual, and knowledge-intensive abilities. The objective of this stage is to establish a strong and well-balanced general expert that serves as the foundation for subsequent post-training.

\begin{table}[t]
\centering
\small
\begin{tabular}{lrr}
\toprule
Data type & Weighted examples & Sampling share \\
\midrule
Code & 1,013,476 & 18.16\% \\
General & 1,169,718 & 20.97\% \\
Math & 2,653,858 & 47.57\% \\
Science & 293,848 & 5.27\% \\
Tool use & 314,481 & 5.64\% \\
Chinese QA & 133,938 & 2.40\% \\
\bottomrule
\end{tabular}
\caption{Stage-1 SFT data mixture. Stage 1 builds a broad general-capability foundation from diverse supervised data.}
\label{tab:sft-stage1-mixture}
\end{table}

After broad capability learning, we further perform a second-stage refinement using a substantially smaller but carefully curated high-quality corpus, as summarized in Table~\ref{tab:sft-stage2-mixture}. Compared with the first stage, this refinement stage places greater emphasis on response quality, reasoning consistency, and instruction-following behaviors. The high-quality supervision further improves the model's alignment and reasoning performance while preserving the broad capabilities established during the first stage.

\begin{table}[t]
\centering
\small
\begin{tabular}{ll}
\toprule
Data type & Role in Stage 2 \\
\midrule
Reasoning & Improve long-form reasoning and general problem solving \\
Tool use & Provide structured tool-use exposure \\
Agent interaction & Support later interactive and tool-mediated training \\
Instruction following & Improve controllability under diverse instructions \\
Multilingual enhancement & Strengthen non-English general capability \\
General SFT mixture & Broaden capability coverage and preserve general performance \\
\bottomrule
\end{tabular}
\caption{Stage-2 SFT data mixture. Stage 2 performs targeted continuation training on top of the Stage-1 checkpoint.}
\label{tab:sft-stage2-mixture}
\end{table}

General SFT is performed sequentially in these two stages, with each stage adopting an independent optimization schedule. Rather than simply extending training on the same corpus, the second-stage refinement serves as a dedicated quality enhancement process, allowing Athena to improve response quality and reasoning consistency while maintaining the broad capabilities acquired during the first stage.

After General SFT, Athena acquires strong general language understanding, instruction-following, reasoning, and coding capabilities, providing a robust initialization for the subsequent General Reinforcement Learning stage.

\subsection{General Reinforcement Learning}

While General SFT establishes strong instruction-following and general reasoning capabilities, supervised learning alone is insufficient for fully optimizing reasoning quality and reasoning efficiency. For practical robotic deployment, a robot foundation model should not only produce correct decisions, but also reach them through concise reasoning under strict latency constraints. Therefore, the second stage of Athena's training pipeline introduces General Reinforcement Learning (General RL) to jointly improve reasoning capability and reasoning efficiency.

To achieve this objective, we optimize the model using a combination of task-specific correctness rewards and a token-budget reward. The correctness rewards are automatically computed for tasks with verifiable answers, including mathematical reasoning, scientific reasoning, and code generation, encouraging accurate and robust problem solving. In addition, we introduce a token-budget reward that explicitly encourages concise reasoning by rewarding solutions that reach correct answers with fewer generated tokens. The model is optimized using Group Relative Policy Optimization (GRPO) \citep{shao2024deepseekmath}, enabling large-scale reinforcement learning on automatically verifiable tasks without human preference annotations.

During training, we constrain the maximum generation budget to 16K output tokens. We experimented with different budget settings and observed that excessively restrictive budgets noticeably degraded performance on complex tasks, whereas substantially larger budgets weakened the effectiveness of the token-budget reward (See Section~\ref{sec:rollout_bugdet}). We therefore adopt a budget of 16K tokens to balance reasoning capability and inference efficiency.

\begin{figure}[t]
    \centering
    \includegraphics[width=0.49\linewidth]{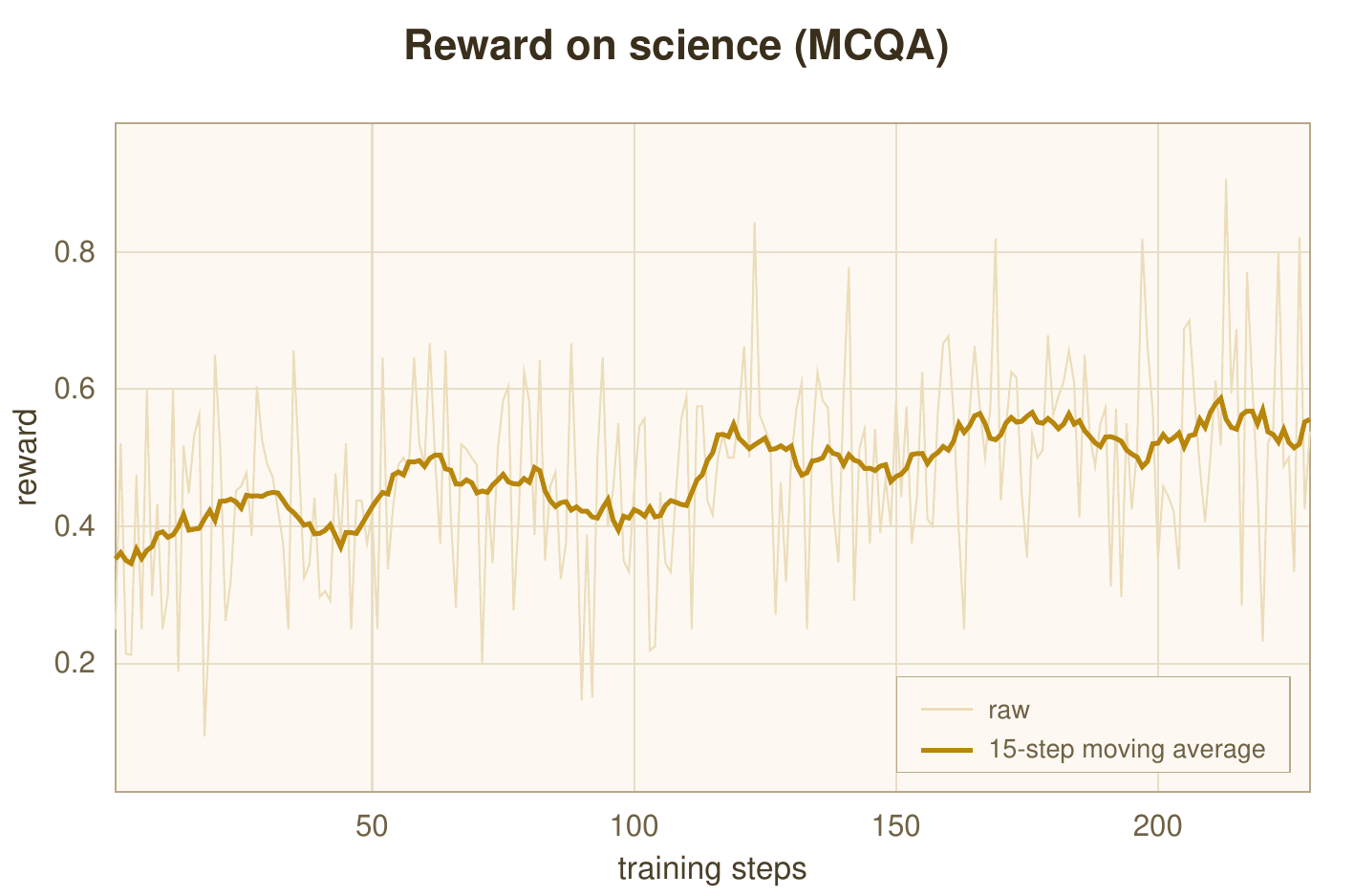}
    \includegraphics[width=0.49\linewidth]{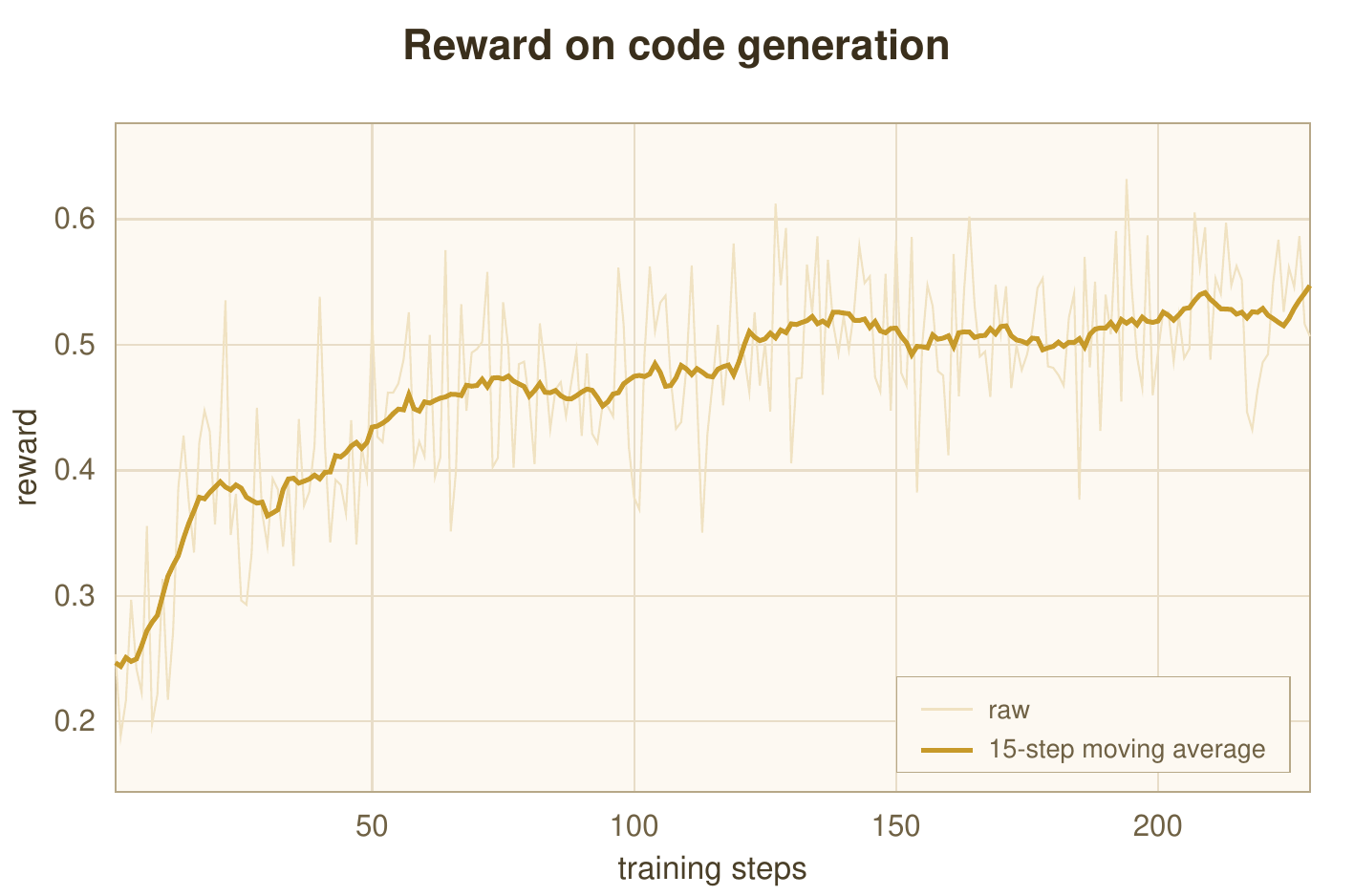}\\
    \includegraphics[width=0.49\linewidth]{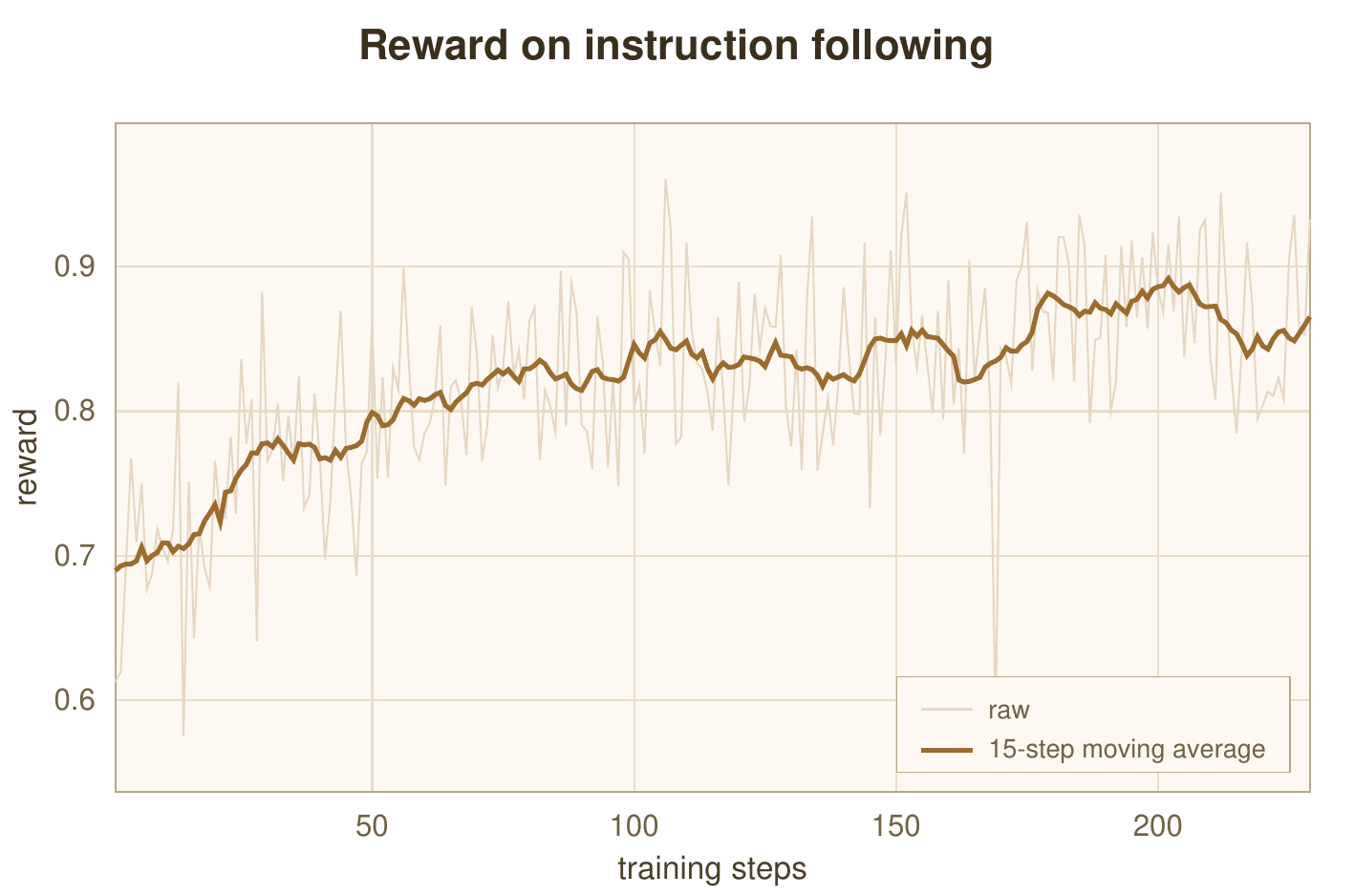}
    \includegraphics[width=0.49\linewidth]{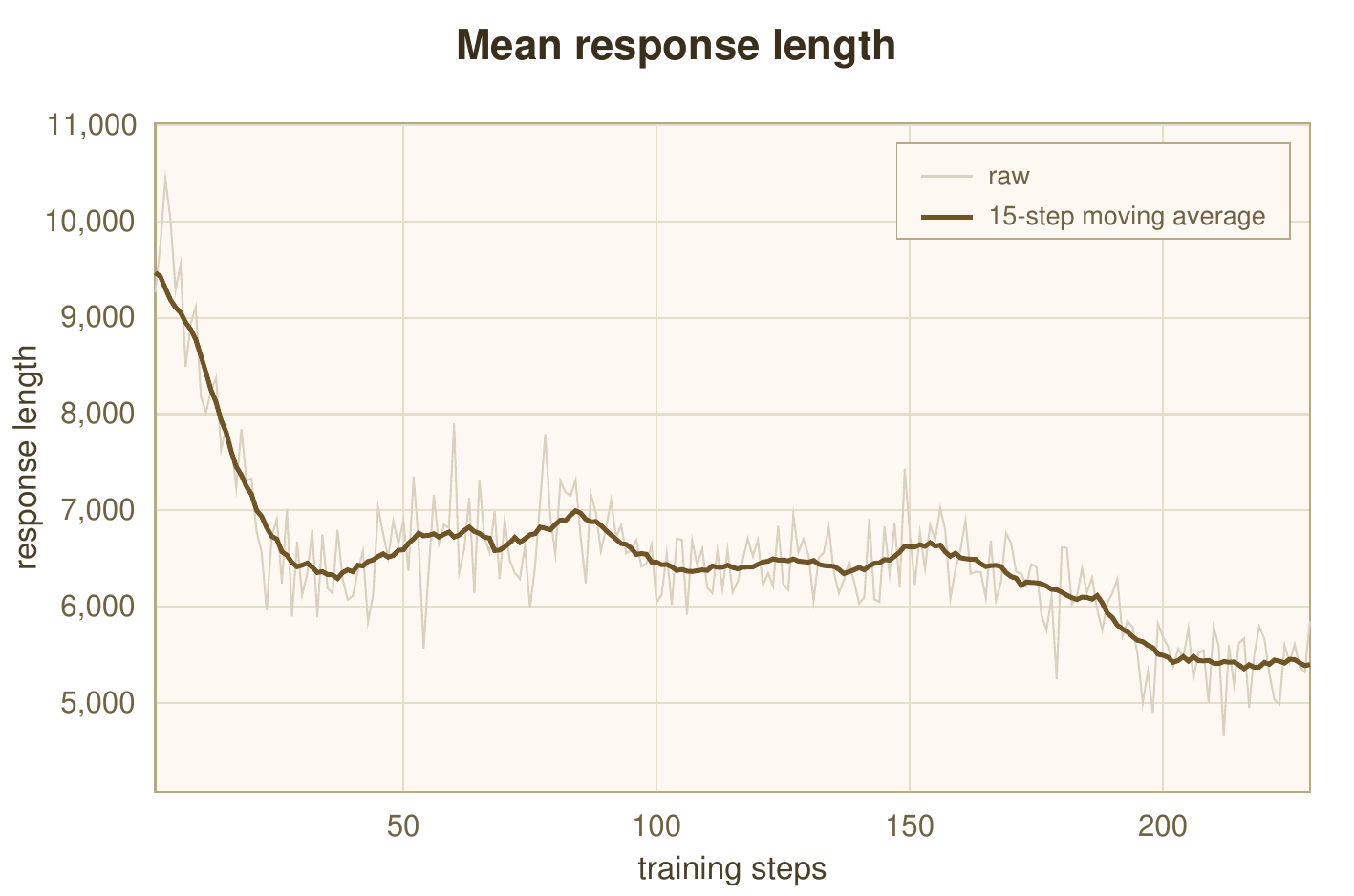}
    \caption{
    Training dynamics during General Reinforcement Learning. The first three subfigures show the reward improvement on representative reasoning tasks throughout reinforcement learning, demonstrating stable optimization and continuous capability enhancement. The bottom-right subfigure illustrates the average generated token consumption during training. Together, these results show that General RL simultaneously improves reasoning capability while encouraging increasingly concise reasoning through the token-budget reward.
    }
    \label{fig:general_rl_curve}
\end{figure}

Figure~\ref{fig:general_rl_curve} illustrates the optimization dynamics during General RL. Across representative reasoning tasks, the reward consistently improves throughout training, indicating stable reinforcement learning and continuous enhancement of reasoning capability. Meanwhile, the average reasoning token consumption gradually decreases as optimization proceeds, demonstrating that the token-budget reward effectively encourages efficient reasoning without compromising task performance.

After General RL, Athena develops a strong and efficient general reasoning expert, providing the foundation for the subsequent embodied expert training stage.


\subsection{Embodied Expert Training}

Although General RL equips Athena with strong general reasoning capabilities and efficient generation behaviors, the resulting model still lacks the embodied knowledge required for real-world robotic interaction. A robot brain model must not only understand and reason about language, but also perceive environments, interpret observations, plan actions, and interact with external systems. To bridge this gap, we introduce an Embodied Expert Training stage that specializes Athena for embodied intelligence while preserving the strong general capabilities acquired in previous stages.

Unlike general language tasks, embodied tasks require continuous interaction with an external environment rather than producing a single response from a fixed input. At each interaction step, the model receives updated observations, reasons about the current state, executes actions, and adapts its behavior according to environmental feedback. We represent all embodied interactions using this unified interaction interface, allowing different scenarios to be trained within a consistent language modeling framework.

\subsubsection{Interactive Scenarios}

\begin{figure}[t]
    \centering   \includegraphics[width=\linewidth]{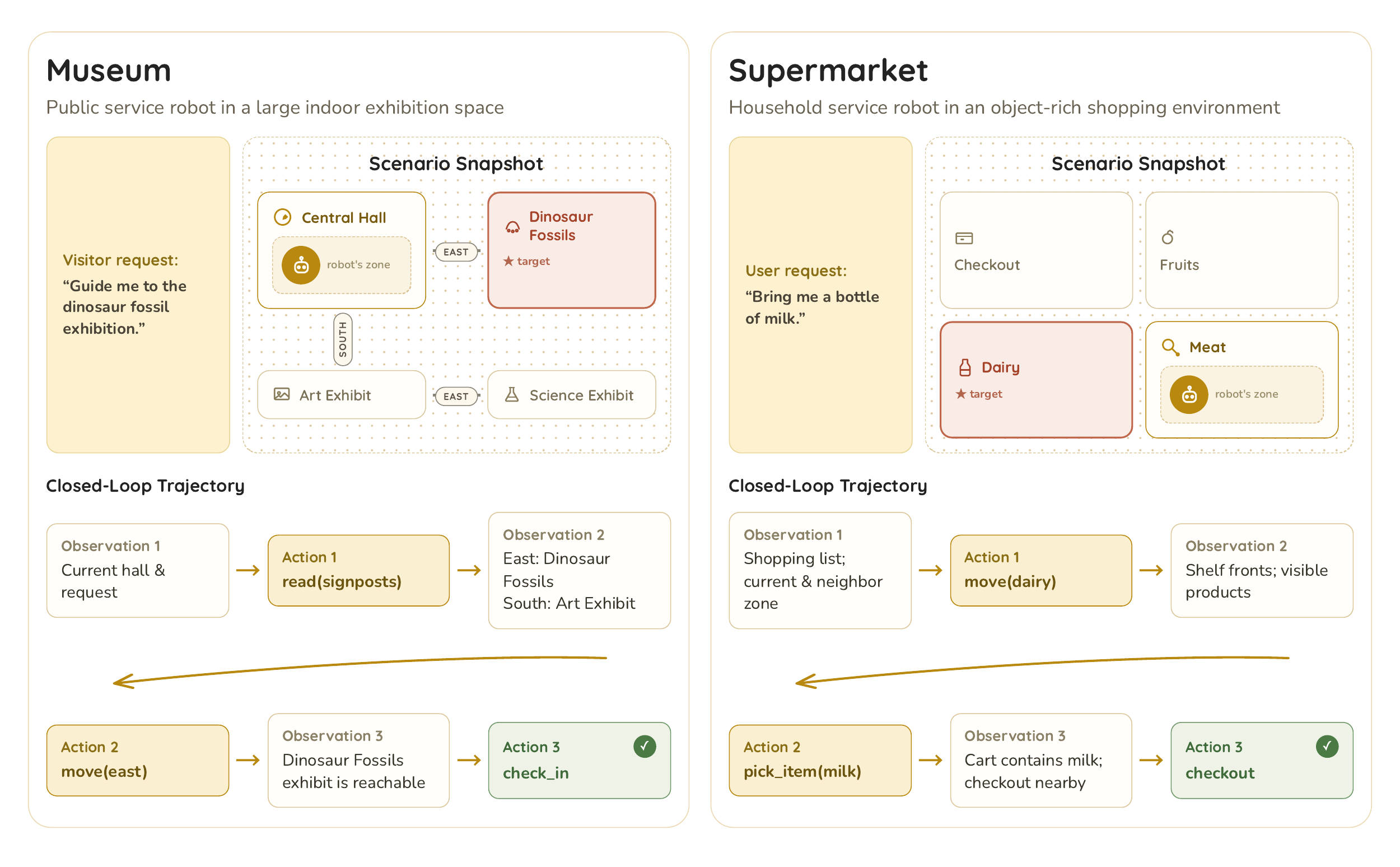}
    \caption{Representative interactive scenarios used for embodied expert training. The Museum and Supermarket environments represent complementary service robotics domains. Despite different task objectives, both environments share the same observation-action interaction loop, providing consistent closed-loop interaction experiences for training Athena.}    \label{fig:interactive_scenarios}
\end{figure}

\begin{table}[!htb]
\centering
\small
\begin{tabular}{p{0.10\linewidth} p{0.24\linewidth} p{0.56\linewidth}}
\toprule
Level & Capability & Representative Tasks \\
\midrule
0 & High-level topology & Task planning, tool use, target search over rooms or zones. \\
1 & Local topology and orientation & Maintaining position under limited local observations and choosing exploratory actions. \\
2 & Continuous position with coarse observations & Estimating distance, approaching interactive objects, and reasoning about local reachability. \\
3 & Continuous heading with view-cone observations & Egocentric localization, turning, and decision making under local visibility constraints. \\
\bottomrule
\end{tabular}
\caption{Progressive embodiment levels adopted in the interactive scenarios for embodied expert training. Higher embodiment levels require increasingly complex environmental interaction, ranging from passive perception to long-horizon task execution, enabling a curriculum that progressively develops Athena's embodied capabilities.}
\label{tab:embodiment_levels}
\end{table}

To support embodied expert training, we construct a collection of executable interactive scenarios that simulate representative service robotics applications. Unlike static instruction-response datasets, these scenarios enable Athena to continuously interact with an external environment through sequential observations and executable actions, providing closed-loop interaction experiences for learning embodied behaviors.

We develop two representative environments that capture complementary service robotics domains, as illustrated in Figure~\ref{fig:interactive_scenarios}. The Museum environment focuses on public service tasks such as navigation, visitor assistance, and human-robot interaction in large indoor spaces. The Supermarket environment emphasizes household service capabilities, including object grounding, mobile manipulation, and task-oriented planning in object-rich environments. Despite different task objectives, both environments share the same observation-action interaction protocol, enabling Athena to learn embodied skills through closed-loop interaction.

\begin{figure}[tp]
    \centering   \includegraphics[width=\linewidth]{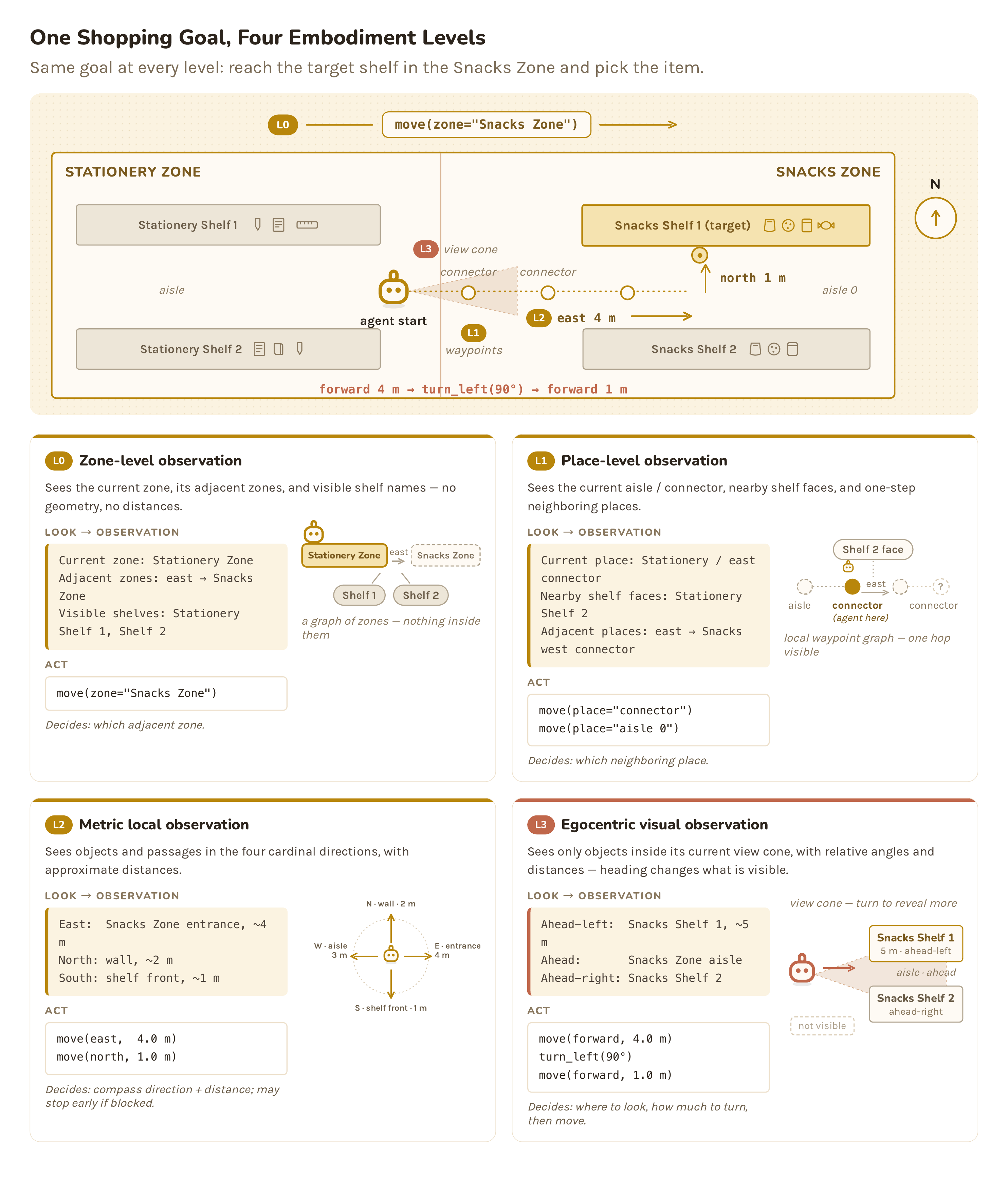}
    \caption{Illustration of the four embodiment levels in the supermarket environment, ranging from zone-level planning to egocentric embodied interaction.}    \label{fig:embody_levels}
\end{figure}

Rather than exposing the model to the most complex interactions from the beginning, tasks within each environment are further organized into progressive embodiment levels according to the complexity of environmental interaction. As summarized in Table~\ref{tab:embodiment_levels}, lower levels focus on perception and grounded action prediction with limited interaction, while higher levels gradually introduce sequential decision making, dynamic environment transitions, and long-horizon task execution. This progressive organization naturally forms a curriculum for embodied expert training, allowing Athena to first acquire fundamental embodied capabilities before gradually mastering increasingly complex interactive behaviors. Figure~\ref{fig:embody_levels} illustrates the four embodiment levels in the supermarket environment, highlighting the progressively richer observations and interaction capabilities across different levels.

Despite the diversity of embodied tasks, all interactive scenarios follow the same observation-action interaction protocol. At each interaction step, Athena receives the current observation from the environment and predicts the next executable action. After the action is executed, the environment returns an updated observation, which becomes the input for the subsequent interaction step. Repeating this observation-action loop enables Athena to continuously learn from environmental feedback while maintaining a unified interaction interface across heterogeneous embodied scenarios.

\subsubsection{Embodied Expert Training Process}

With executable interactive scenarios in place, Athena is further specialized through embodied expert training. While General SFT and General RL primarily improve language understanding, reasoning, and response behaviors, this stage focuses on acquiring embodied interaction capabilities through continuous interaction with executable environments. By repeatedly observing the environment, executing actions, and receiving environmental feedback, Athena gradually learns navigation, grounded action prediction, sequential decision making, and long-horizon task execution. Training begins with a 4K-step supervised warm-up on collected interaction trajectories, followed by GRPO-based reinforcement learning to further optimize embodied interaction policies through environmental feedback.

Training trajectories are collected from the interactive scenarios described in the previous section. Depending on the embodiment level, trajectories cover progressively more challenging interaction scenarios, ranging from basic grounded action execution to long-horizon decision making with dynamic environment transitions. All interaction trajectories are represented using a unified observation-action format. During the supervised warm-up stage, these trajectories provide demonstration supervision, while the subsequent GRPO stage further optimizes the policy using executable environment rewards, enabling Athena to learn consistent interaction behaviors across heterogeneous embodied tasks. Rewards are accumulated from executable environment scores: Museum rewards target-visit progress and completion, while Supermarket rewards cart correctness and checkout success. Rollouts stop on task completion, the 100-step episode limit, or one counted tool-protocol violation; under the relaxed truncation policy, malformed or missing tool calls truncate the rollout, whereas unknown tools, locked tools, and executable skill failures remain recoverable.

\begin{figure}[t]
\centering
\begin{subfigure}{0.48\linewidth}
\centering
\includegraphics[width=\linewidth]{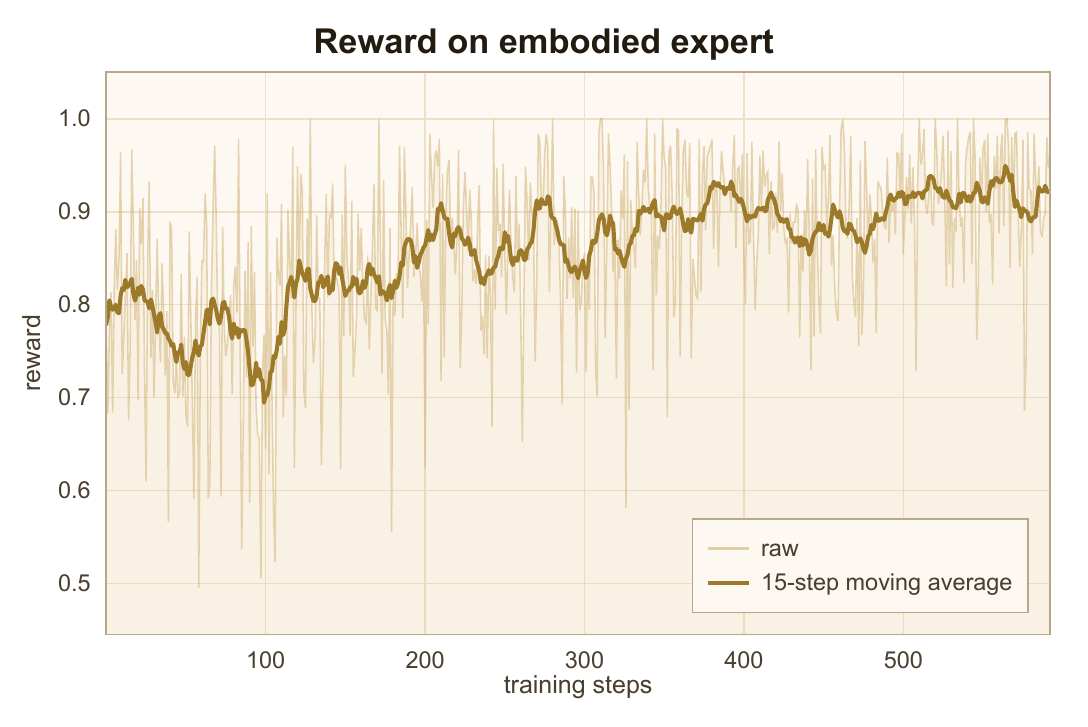}
\caption{RL training reward.}
\label{fig:rl_reward_curve}
\end{subfigure}
\hfill
\begin{subfigure}{0.48\linewidth}
\centering
\includegraphics[width=\linewidth]{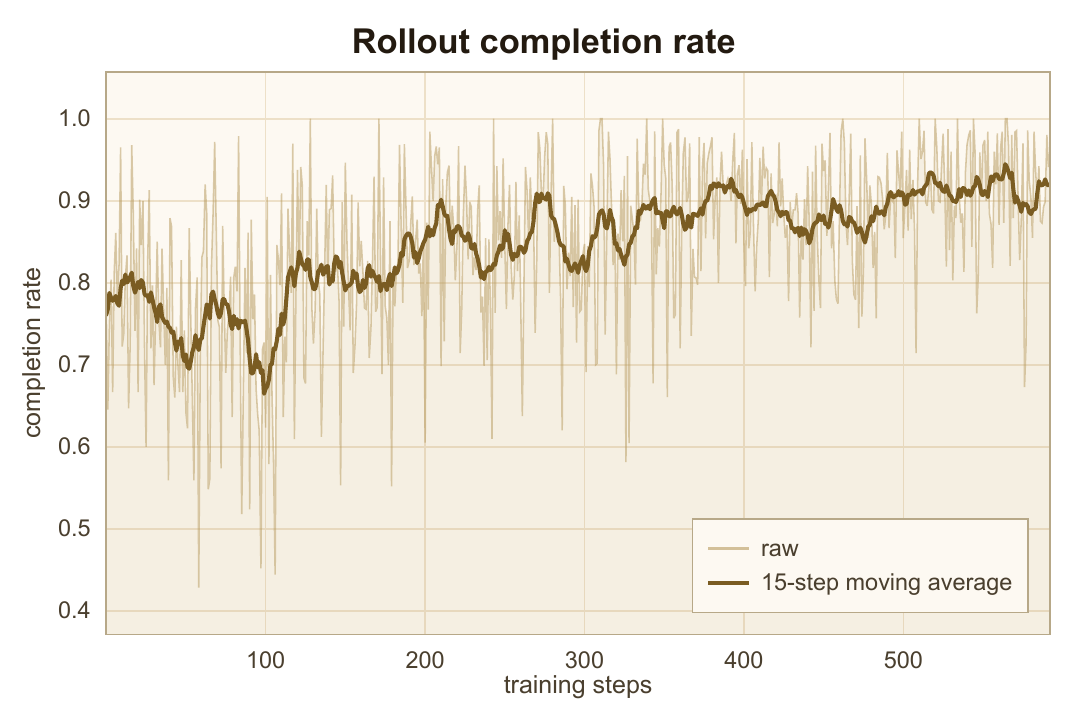}
\caption{Rollout completion rate.}
\label{fig:rl_completion_curve}
\end{subfigure}
\caption{
Training dynamics during embodied expert training.
Both the average interaction reward and the roll-out completion rate steadily increase, indicating progressively improved interaction quality and stronger task-solving ability throughout training.
}
\label{fig:embodied_training}
\end{figure}

Figure~\ref{fig:embodied_training} presents the training dynamics during embodied expert training. As training progresses, both the average interaction reward and the task completion rate steadily increase, indicating that Athena continuously improves its decision-making quality and completes more tasks within executable environments. Together, these trends show that Athena effectively acquires embodied interaction capabilities through progressive training.

The resulting embodied expert model serves as a specialized branch for embodied intelligence. In the next stage, it is merged with the general model to obtain the final Athena model, preserving both strong general language capabilities and robust embodied interaction skills.

\subsection{Model Merging}
\label{sec:model-merge}

The training stages described above produce multiple specialized checkpoints with complementary strengths. General supervised fine-tuning establishes broad language capability, general reinforcement learning improves reasoning quality and efficiency, while embodied expert training specializes the model for different interactive environments. Rather than selecting a single checkpoint or performing additional joint optimization, we combine these capabilities through a lineage-aware, two-stage parameter-space merge to obtain the final Athena-Brain-8B.

In the first stage, we consolidate the domain-specialized RL experts that share the same Athena-SFT initialization. We define their task vectors relative to this common SFT anchor and apply TIES~\cite{yadav2023ties} to merge the resulting updates. Because these experts share a training lineage, coordinate-wise sign agreement reflects agreement among comparable post-training updates; TIES therefore preserves salient, aligned modifications while suppressing weak or conflicting changes.

In the second stage, we introduce a complementary SFT checkpoint from a different training lineage through conservative, low-weight linear interpolation with the first-stage model. We keep this checkpoint outside the TIES voting pool because its displacement relative to the Athena-SFT anchor contains lineage-level as well as task-specific differences, making coordinate-wise conflict resolution poorly grounded. The resulting linear correction transfers complementary capability while preserving the local parameter geometry established by the same-origin RL merge.

Our objective is not to maximize performance on any single benchmark, but to obtain a unified robot-brain model that simultaneously maintains strong general language capability, efficient reasoning, and robust embodied interaction skills. Accordingly, the merging coefficients are selected based on overall capability balance across both general and embodied evaluations, rather than optimizing for a specific task or environment. The resulting merged checkpoint, denoted as \textbf{Athena-Brain-8B}, serves as the final model evaluated throughout the remainder of this report.

%% file: main/evaluation.tex
\section{Evaluation}
\label{sec:evaluation}

\paragraph{Evaluation Overview.}
We evaluate Athena-Brain-8B from two complementary perspectives. First, we assess its general capability on a broad benchmark suite covering mathematics, coding, reasoning, general tasks, and function calling, and further examine its reasoning efficiency in terms of both performance and token usage. This is particularly important for a robot-brain model, which must produce high-quality decisions under practical latency and resource constraints. Second, we evaluate embodied interaction capability in text-based interactive environments, where the model must repeatedly perceive observations, invoke actions, and adapt to environmental feedback over long-horizon interaction loops. We conduct in-domain evaluation on the Museum Tour and Supermarket Shopping environments under progressively increasing embodiment levels, and additionally report zero-shot transfer results on the unseen ALFWorld split to examine whether the acquired interaction skills generalize beyond the training environments.

We compare Athena with representative open-source language models covering both general-purpose and embodied settings. Qwen3-8B~\citep{yang2025qwen3technicalreport} serves as the primary general-purpose baseline, sharing the same base model as Athena and providing a direct comparison for the effectiveness of our post-training pipeline. Athena-SFT, the intermediate supervised fine-tuned model, is included to quantify the contribution of the subsequent reinforcement learning stages. We further compare against recent compact 7B--8B embodied or edge-oriented language models, including MiniCPM4.1-8B~\citep{minicpm4}, MiMo-Embodied-7B~\citep{hao2025mimoembodiedxembodiedfoundationmodel}, and RynnBrain-8B~\citep{dang2026rynnbrain}, which represent current approaches to compact embodied intelligence. 

For the in-domain embodied interaction benchmarks, we additionally evaluate representative frontier LLMs, including GPT-5.5~\citep{openai2026gpt55}, Gemini-3.1-Pro~\citep{googledeepmind2026gemini31pro}, DeepSeek-V4-Pro~\citep{deepseekai2026deepseekv4}, Kimi K2.6~\citep{moonshotai2026kimik26}, Qwen3-Max~\citep{qwen2025qwen3max}, and Qwen3.7-Max~\citep{qwen2026qwen37max}, to assess how a compact embodied language model compares with today's leading large-scale models on embodied interaction tasks.
 
\subsection{General Capability}
\label{sec:general_capability}

\begin{table*}[t]
\centering
\small
\setlength{\tabcolsep}{4pt}
\resizebox{\textwidth}{!}{%
\begin{tabular}{lcccccc}
\toprule
Dataset & Athena-Brain-8B & Athena-SFT & RynnBrain-8B & MiniCPM4.1-8B & Mimo-Embodied-7B & Qwen3-8B \\
\midrule
\multicolumn{7}{l}{\textit{Math}} \\
AIME 2024 & \textbf{76.67} & 73.75 & 2.92 & 67.08 & 62.08 & 75.83 \\
AIME 2025 & \textbf{70.83} & 64.58 & 1.25 & 59.58 & 50.83 & 68.75 \\
BeyondAIME & \textbf{51.00} & 49.00 & 4.00 & 31.00 & 25.00 & 43.00 \\
OlympiadBench & \textbf{79.70} & 76.30 & 16.15 & 72.30 & 70.52 & 76.89 \\
\textbf{Math Average} & \textbf{69.55} & 65.91 & 6.08 & 57.49 & 52.11 & 66.12 \\

\midrule
\multicolumn{7}{l}{\textit{Code \& Reasoning}} \\
LiveCodeBench v5 & \textbf{58.43} & 53.01 & 23.49 & 53.61 & 40.96 & 53.01 \\
LiveCodeBench v6 & \textbf{54.63} & 48.90 & 25.33 & 46.92 & 37.00 & 53.08 \\
ZebraLogic & 76.70 & 74.00 & 3.10 & 45.30 & 58.70 & \textbf{85.60} \\
\textbf{Code \& Reasoning Average} & 63.25 & 58.64 & 17.31 & 48.61 & 45.56 & \textbf{63.90} \\

\midrule
\multicolumn{7}{l}{\textit{General Tasks}} \\
Arena-Hard v2 & \textbf{48.21} & 43.26 & 4.16 & 18.87 & 30.58 & 37.73 \\
GPQA-Diamond & 59.60 & 59.09 & 40.91 & 31.82 & 51.01 & \textbf{61.62} \\
IFEval & \textbf{86.32} & 78.19 & 73.01 & 73.20 & 76.52 & 85.40 \\
MMLU-Redux & 86.47 & 85.86 & 74.25 & 86.16 & 85.26 & \textbf{86.74} \\
SuperGPQA & \textbf{46.64} & 41.20 & 27.86 & 20.83 & 39.55 & 40.31 \\
LiveBench & 70.80 & 68.60 & 44.50 & 55.60 & 60.70 & \textbf{75.80} \\
Humanity's Last Exam & 12.18 & 14.60 & 5.04 & \textbf{26.37} & 10.61 & 8.46 \\
\textbf{General Tasks Average} & \textbf{58.60} & 55.83 & 38.53 & 44.69 & 50.61 & 56.58 \\

\midrule
\multicolumn{7}{l}{\textit{Function Calling}} \\
BFCL v1 & 81.51 & 82.66 & 82.01 & 17.27 & 70.79 & \textbf{84.46} \\
BFCL v2 & 76.94 & \textbf{79.21} & 71.48 & 39.27 & 59.53 & 77.52 \\
BFCL v3 & \textbf{43.12} & 39.88 & 18.62 & 0.00 & 6.88 & 36.88 \\
TAU2 Airline & 46.00 & \textbf{54.00} & 26.00 & 36.00 & 22.00 & 32.00 \\
TAU2 Retail & \textbf{45.61} & 40.35 & 5.26 & 6.14 & 24.56 & 32.46 \\
TAU2 Telecom & \textbf{45.61} & 44.74 & 10.53 & 6.14 & 11.40 & 18.42 \\
\textbf{Function Calling Average} & 56.47 & \textbf{56.81} & 35.65 & 17.47 & 32.53 & 46.96 \\

\midrule
\textbf{Overall Average} & \textbf{60.85} & 58.56 & 27.99 & 39.67 & 44.73 & 56.70 \\
\bottomrule
\end{tabular}%
}
\caption{General capability evaluation of Athena-Brain-8B against representative 8B models. We compare the final Athena-Brain-8B with Athena-SFT, Qwen3-8B, MiniCPM4.1-8B, Mimo-Embodied-7B, and RynnBrain-8B. Results are reported on individual benchmarks together with averaged scores over mathematics, code and reasoning, general tasks, and function calling.}
\label{tab:sft-full-results}
\end{table*}

To assess the general capability of Athena-Brain-8B, we evaluate the model on a diverse collection of public benchmarks spanning mathematical reasoning, coding, scientific reasoning, general language understanding, and function calling. The benchmark suite is organized into the following categories:
\begin{itemize}[leftmargin=*]
    \item \textbf{Mathematical Reasoning:} AIME 2024, AIME 2025~\citep{aops_aime}, BeyondAIME~\citep{bytedance_seed_2025_beyondaime}, and OlympiadBench~\citep{he2024olympiadbench}.
    
    \item \textbf{Coding and Scientific Reasoning:} LiveCodeBench~\citep{jain2025livecodebench} and ZebraLogic~\citep{zebralogic2024}.
    
    \item \textbf{General Capability:} Arena-Hard-v2~\citep{li2024crowdsourced}, GPQA-Diamond~\citep{rein2024gpqa}, IFEval~\citep{zhou2023ifeval}, MMLU-Redux~\citep{gema2025mmluredux}, SuperGPQA~\citep{du2025supergpqa}, LiveBench~\citep{white2025livebench} and Humanity's Last Exam~\citep{phan2025hle}.
    
    \item \textbf{Function Calling:} BFCL~\citep{patil2025bfcl} and $\tau^2$-Bench~\citep{barres2025tau2}.
\end{itemize}

\paragraph{Evaluation Protocol.}

We compare Athena-Brain-8B against representative open-weight language models spanning both general-purpose and reasoning-oriented models of comparable scale. All evaluations are performed on the DeepInsight evaluation platform~\cite{li2026deepinsightunifiedevaluationinfrastructure} to ensure a unified and reproducible evaluation protocol across benchmarks. Unless otherwise specified, all models are evaluated under the same inference configuration to ensure a fair comparison.

All models are evaluated with thinking mode enabled. We set the maximum generation length to 40,960 tokens. For models trained with a native context length of 32,768 tokens, we apply YaRN-based context extension to match the evaluation budget. Decoding is performed with temperature $0.6$, top-$p$ $0.95$, and top-$k$ $20$, following a unified reasoning-mode evaluation protocol across all benchmarks and model variants.

For benchmarks with relatively small evaluation sets, stochastic decoding may introduce noticeable variance. Therefore, we repeat the evaluation eight times on AIME 2024 and AIME 2025 and report the averaged results. For all other benchmarks, we report the standard single-run results under the shared evaluation configuration.

\paragraph{Overall Results.}

Table~\ref{tab:sft-full-results} summarizes the performance of Athena-Brain-8B on a diverse suite of general benchmarks covering mathematics, coding, reasoning, knowledge, instruction following, and function calling. Overall, Athena-Brain-8B demonstrates strong and well-balanced general capability across a wide range of tasks while maintaining an efficient 8B model scale.

Compared with the intermediate Athena-SFT checkpoint, reinforcement learning consistently improves reasoning-intensive tasks, particularly mathematical reasoning and code generation, while preserving strong performance on knowledge-intensive and tool-use benchmarks. These results indicate that the general RL stage effectively enhances reasoning ability without sacrificing the broad language capabilities established during supervised fine-tuning.

\paragraph{Reasoning Efficiency.}

\begin{figure}[t]
    \centering
    \includegraphics[width=1\linewidth]{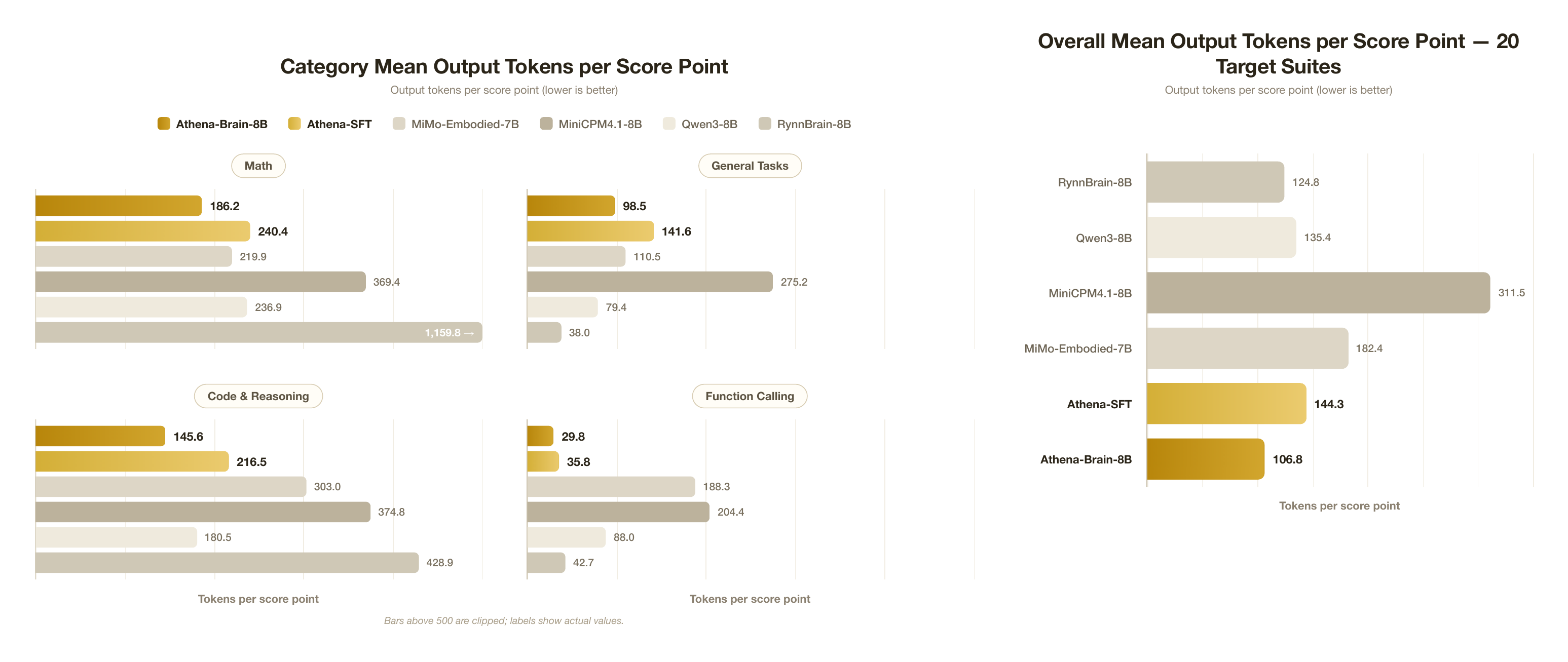}
    \caption{
    Average generated tokens of different models across representative benchmark categories.
    Athena-Brain consistently generates shorter responses than Athena-SFT while maintaining stronger capability.
    }
    \label{fig:length_comparison}
\end{figure}

For embodied agents, benchmark accuracy alone is not sufficient. During long-horizon interaction, the model repeatedly alternates between perception, reasoning, and action generation. Excessively long reasoning chains therefore accumulate inference latency and token cost, reducing interaction efficiency and limiting practical deployment. Consequently, an ideal robot brain should achieve strong reasoning performance while minimizing unnecessary token generation.

To evaluate the trade-off between reasoning capability and inference efficiency, we measure \textbf{token cost} as the ratio of average generated tokens to average benchmark score (lower is better). Figure~\ref{fig:length_comparison} compares representative reasoning models in terms of benchmark performance and average generated tokens. Athena-Brain-8B achieves a favorable balance between reasoning capability and inference efficiency, reaching competitive benchmark performance while requiring substantially fewer generated tokens than long-chain reasoning models. This behavior reflects the objective of our general RL stage, which encourages concise yet effective reasoning rather than maximizing reasoning length.

The resulting efficiency is particularly valuable for embodied interaction, where models must make frequent decisions under latency constraints. By reducing reasoning overhead without compromising capability, Athena-Brain-8B provides a practical foundation for responsive and efficient robot control.

\subsection{Embodied Interaction Capability}

We evaluate Athena-Brain-8B on text-based interactive environments that require repeated perception, action, and feedback-driven adaptation over long-horizon episodes. In addition to in-domain evaluation on the Museum Tour and Supermarket Shopping environments, we also assess zero-shot transfer on ALFWorld~\citep{shridhar2021alfworld} to probe generalization beyond the training environments. To better understand the source of the observed gains, we further introduce two intermediate probes that isolate goal-oriented completion and spatial recovery under partial observability.

\subsubsection{Evaluation Protocol}

We evaluate Athena-Brain-8B in both in-domain and out-of-domain text-based interactive environments under a unified interaction protocol. At each interaction step, the model receives the complete dialogue history, including the task instruction, previous observations, previous actions, execution feedback from the environment, and the available tool interface. The model is required to generate a single structured JSON tool call, which is executed by the simulator to update the environment state and return the next observation.

To ensure a fair comparison, all models are evaluated under the same decoding configuration, with temperature set to $0.6$, top-$p$ to $0.95$, and top-$k$ to $20$. For proprietary API models, we use their standard inference mode while allowing the model to autonomously determine the amount of reasoning required for each interaction, reflecting practical embodied deployment where excessive deliberation may increase interaction latency. For open-weight and API models, we follow their recommended inference settings and additionally evaluate both the standard and thinking variants when available (e.g., Qwen3-8B), providing a comprehensive comparison of different reasoning behaviors. This protocol also enables us to analyze the relationship between interaction performance and reasoning efficiency by comparing both task success and the number of generated tokens across different inference behaviors.

For in-domain evaluation, we use the Museum Tour and Supermarket Shopping environments introduced in Section~3.3.1. Both environments are evaluated under four embodiment levels ($E0$--$E3$), which progressively increase spatial and interaction complexity. The evaluation set contains 100 museum instances and 120 supermarket instances, with a maximum interaction budget of 100 interaction steps per episode. A museum episode is considered successful only if the agent reaches the designated target and completes the tour, while a supermarket episode requires checking out with a shopping cart satisfying all task constraints.

For out-of-domain evaluation, we conduct zero-shot evaluation on the unseen split of ALFWorld without additional fine-tuning. Models interact with the environment solely through textual observations and executable actions under the same interaction protocol. We report Success Rate (SR) as the primary evaluation metric for all embodied benchmarks.

\subsubsection{In-domain Interactive Performance}
\label{sec:in_domain_interactive}

\begin{table}[t]
\centering
\scriptsize
\setlength{\tabcolsep}{3.5pt}
\begin{tabular}{lccccccccccc}
\toprule
\multirow{2}{*}{Model}
& \multicolumn{4}{c}{Museum}
& \multicolumn{4}{c}{Supermarket}
& \multirow{2}{*}{Museum Avg.}
& \multirow{2}{*}{Supermarket Avg.}
& \multirow{2}{*}{Overall} \\
\cmidrule(lr){2-5} \cmidrule(lr){6-9}
& $E0$ & $E1$ & $E2$ & $E3$
& $E0$ & $E1$ & $E2$ & $E3$
& & & \\
\midrule
GPT-5.5 & 97.0 & 78.0 & 90.0 & 55.0 & 95.0 & 94.2 & 81.7 & 45.8 & 80.0 & 79.2 & 79.6 \\
Gemini-3.1-Pro & 95.0 & 54.0 & 84.0 & 38.0 & 95.8 & 85.8 & 80.8 & 35.0 & 67.8 & 74.4 & 71.1 \\
DeepSeek-V4-Pro & 77.0 & 32.0 & 60.0 & 19.0 & 95.0 & 75.8 & 53.3 & 14.2 & 47.0 & 59.6 & 53.3 \\
Kimi-K2.6 & 69.0 & 35.0 & 64.0 & 21.0 & 95.8 & 71.7 & 51.7 & 10.8 & 47.3 & 57.5 & 52.4 \\
Qwen3.7-Max & 60.0 & 60.0 & 61.0 & 28.0 & 93.3 & 78.3 & 38.3 & 14.2 & 52.3 & 56.0 & 54.1 \\
Qwen3-Max & 49.0 & 43.0 & 48.0 & 26.0 & 89.2 & 59.2 & 21.7 & 10.0 & 41.5 & 45.0 & 43.3 \\
\midrule
MiniCPM4.1-8B & 0.0 & 0.0 & 0.0 & 0.0 & 2.5 & 0.0 & 0.0 & 0.0 & 0.0 & 0.63 & 0.32 \\
MiMo-Embodied-7B & 16.0 & 6.0 & 4.0 & 2.0 & 1.67 & 0.0 & 0.0 & 0.0 & 7.0 & 0.42 & 3.71 \\
RynnBrain-8B & 21.0 & 9.0 & 3.0 & 0.0 & 0.0 & 0.0 & 0.0 & 0.0 & 8.25 & 0.0 & 4.13 \\
Qwen3-8B-NonThinking      & 16.0 & 14.0 & 6.0 & 2.0 & 12.5 & 5.0 & 0.83 & 0.0 & 9.5 & 4.58 & 7.04 \\
Qwen3-8B-Thinking      & 46.0 & 11.0 & 17.0 & 3.0 & 24.17 & 10.83 & 0.0 & 0.0 & 19.25 & 8.75 & 14.0 \\
Athena-SFT    & 16.0 & 1.0 & 3.0 & 0.0 & 28.33 & 6.67 & 0.83 & 0.0 & 5.0 & 8.96 & 6.98 \\
\textbf{Athena-Brain-8B}
              & 93.0 & 47.0 & 79.0 & 25.0
              & 78.33 & 73.33 & 57.5 & 15.0
              & 61.0 & 56.04 & 58.52 \\
\bottomrule
\end{tabular}
\caption{
In-domain interactive evaluation of representative frontier API models and open-weight 8B models. Results are reported as success rate (\%) on the Museum Tour and Supermarket Shopping environments. Museum Avg. and Supermarket Avg. are averaged over the four embodiment levels ($E0$--$E3$), and Overall is averaged over all eight environment-level settings.
}
\label{tab:interactive_indomain}
\end{table}

\begin{table}[t]
\centering
\small
\begin{tabular}{lccc}
\toprule
\textbf{Model} & \textbf{Avg. Tokens $\downarrow$} & \textbf{Overall Score $\uparrow$} & \textbf{Token Cost $\downarrow$} \\
\midrule
MiniCPM4.1-8B & 79.38 & 0.32 & 248.08 \\
MiMo-Embodied-7B & 150.96 & 3.71 & 40.69 \\
RynnBrain-8B & 144.09 & 4.13 & 34.89 \\
Qwen3-8B-NonThinking & 101.35 & 7.04 & 14.40 \\
Qwen3-8B-Thinking & 515.38 & 14.00 & 36.81 \\
Athena-SFT & 83.49 & 6.98 & 11.96 \\
\textbf{Athena-Brain-8B} & \textbf{24.25} & \textbf{58.52} & \textbf{0.41} \\
\bottomrule
\end{tabular}
\caption{Reasoning efficiency comparison on the in-domain embodied interaction benchmark. Token Cost is computed as the average generated tokens divided by the overall score, measuring the token expenditure required per unit of task performance. Lower values indicate better reasoning efficiency.}
\label{tab:interactive_efficiency}
\end{table}

Table~\ref{tab:interactive_indomain} reports the in-domain interactive performance of representative frontier API models and open-weight 7B/8B models on the Museum Tour and Supermarket Shopping environments. As the embodiment level increases from $E0$ to $E3$, the success rate consistently decreases across all models, indicating that embodied interaction becomes substantially more challenging under stronger partial observability and longer-horizon decision making. Specifically, Athena-Brain-8B is trained on the Museum and Supermarket environment families, while external baselines are evaluated zero-shot without environment-specific training.

Among the 7B/8B models, Athena-Brain-8B achieves the strongest overall performance by a large margin, reaching an overall success rate of 58.52\%. Compared with the intermediate Athena-SFT checkpoint (6.98\%), Athena-Brain-8B improves performance consistently across both environments, achieving average success rates of 61.0\% on Museum Tour and 56.04\% on Supermarket Shopping. It also substantially outperforms existing compact embodied models, including Qwen3-8B, MiMo-Embodied-7B, MiniCPM4.1-8B, and RynnBrain-8B. These results demonstrate that Athena-Brain-8B effectively acquires robust closed-loop interaction capability while maintaining the compact 8B model size. Although frontier API models still achieve higher absolute performance, Athena-Brain-8B substantially narrows the gap while remaining a compact 8B model.

Table~\ref{tab:interactive_efficiency} further compares the reasoning efficiency of open-weight models. Enabling extended reasoning improves the performance of Qwen3-8B, but increases the average generation length from 101.35 to 515.38 tokens per interaction, resulting in a higher Token Cost (14.40$\rightarrow$36.81). In contrast, Athena-Brain-8B achieves both the highest interaction performance and the lowest generation cost, requiring only 24.25 generated tokens per interaction on average and a Token Cost of 0.41. These results suggest that, for compact embodied language models, improving reasoning efficiency is more beneficial than simply producing longer reasoning traces.

Overall, these results demonstrate that Athena-Brain-8B effectively bridges the gap between compact open-weight models and frontier API models on embodied interaction. By jointly improving interaction capability and reasoning efficiency, Athena-Brain-8B provides a practical foundation for latency-sensitive embodied applications.

\subsubsection{Out-of-domain Generalization}
\label{sec:out_of_domain}

\begin{table}[t]
\centering
\small
\begin{tabular}{lc}
\toprule
Model & Success Rate (\%) \\
\midrule
Athena-SFT & 11.94 \\
\textbf{Athena-Brain-8B} & \textbf{36.57} \\
\midrule
Qwen3-8B-NonThinking & 20.15 \\
MiniCPM4.1-8B & 40.30 \\
MiMo-Embodied-7B & 20.90 \\
RynnBrain-8B & 11.19 \\
\bottomrule
\end{tabular}
\caption{Out-of-domain transfer evaluation on the unseen split of ALFWorld. Athena models are evaluated zero-shot without any ALFWorld-specific training or fine-tuning.}
\label{tab:interactive_out_domain}
\end{table}

To evaluate whether the embodied interaction capability learned during training generalizes beyond the training environments, we conduct zero-shot evaluation on the unseen ALFWorld benchmark. Since ALFWorld differs substantially from the Museum Tour and Supermarket Shopping environments in task semantics, object distributions, and interaction dynamics, successful transfer requires the model to acquire generalizable interaction strategies rather than environment-specific behaviors.

Table~\ref{tab:interactive_out_domain} summarizes the zero-shot transfer results. Although Athena-Brain-8B is trained exclusively on the Museum Tour and Supermarket Shopping environments, it improves the success rate from 11.94\% to 36.57\% over the Athena-SFT checkpoint without any ALFWorld-specific training or fine-tuning. This substantial improvement indicates that the embodied post-training stage equips the model with interaction capabilities that transfer effectively beyond the environments observed during training.

Compared with representative open-weight 8B models, Athena-Brain-8B achieves competitive zero-shot transfer performance despite being trained with only two embodied environments. While the absolute success rate still leaves room for improvement, the substantial gain over Athena-SFT demonstrates that the embodied post-training stage effectively improves transfer to previously unseen environments. Since ALFWorld is used solely as an unseen evaluation benchmark rather than an optimization target, these results provide evidence that the interaction capabilities learned from our limited embodied training data are transferable beyond the training domains.

\subsubsection{Intermediate Interactive Probes}

\begin{table}[t]
\centering
\small
\begin{tabular}{lcccc}
\toprule
Probe & Target & Domains & Horizon & Scoring \\
\midrule
State-goal (GSR) & future state & both & $H_g=5$ calls & predicate \\
State-return (RLSR) & past location & both & $H_r=10$ calls & location \\
\bottomrule
\end{tabular}
\caption{Intermediate interactive evaluation probes. Both probe sets are sampled
from successful reference trajectories and scored by executing model tool calls
from the restored environment state.}
\label{tab:interactive_intermediate_metrics}
\end{table}

\begin{figure}[t]
\centering
\includegraphics[width=\linewidth]{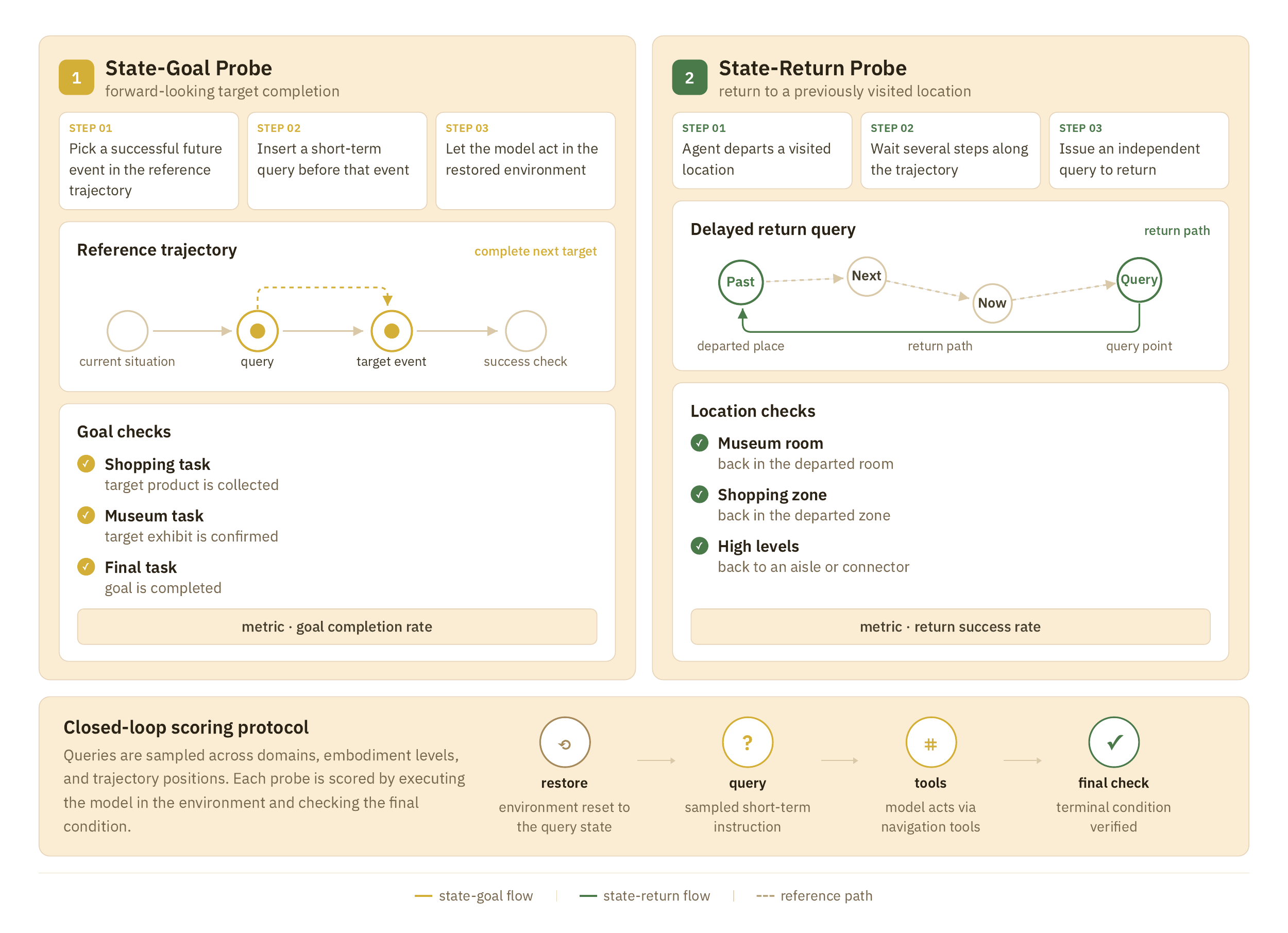}
\caption{
Two intermediate probes for embodied interaction. State-Goal evaluates near-future goal completion, while State-Return evaluates return to a previously visited location under partial observability.
}
\label{fig:intermediate_metrics}
\end{figure}

Final task success provides an end-to-end measure of embodied interaction capability, but it offers limited insight into which underlying cognitive abilities have improved. To better understand the effects of embodied expert training, we introduce two intermediate interactive probes that isolate complementary capabilities required for long-horizon interaction: \textbf{State-Goal}, which evaluates short-horizon goal completion, and \textbf{State-Return}, which evaluates the ability to recover previously visited locations under partial observability. Figure~\ref{fig:intermediate_metrics} illustrates the two probes, while Table~\ref{tab:interactive_intermediate_metrics} summarizes their evaluation settings.

The probe set is automatically constructed from successful interaction trajectories generated by Gemini~3.1~Pro on the \emph{test split} of our in-domain evaluation environments. Using the held-out evaluation set ensures that the probes remain completely disjoint from the embodied training data while providing high-quality reference trajectories for probe generation. For each sampled state along a successful trajectory, the simulator restores the corresponding environment state and inserts an intermediate query. The evaluated model then continues interaction from the restored state by issuing executable tool calls, and success is determined entirely by the resulting environment state after execution.

State-Goal measures whether the model can complete a specified near-future objective within a limited interaction budget, while State-Return evaluates whether the model can successfully navigate back to a previously visited location after moving away from it. Each probe contains 400 evaluation instances, uniformly sampled across the two in-domain environments and all four embodiment levels.

\begin{table}[t]
\centering
\scriptsize
\setlength{\tabcolsep}{2.8pt}
\begin{tabular}{lcccccccccc}
\toprule
\multirow{2}{*}{Model}
& \multicolumn{5}{c}{State-Goal (GSR)}
& \multicolumn{5}{c}{State-Return (RLSR)} \\
\cmidrule(lr){2-6} \cmidrule(lr){7-11}
& $E0$ & $E1$ & $E2$ & $E3$ & Overall
& $E0$ & $E1$ & $E2$ & $E3$ & Overall \\
\midrule
GPT-5.5 & 72.1 & 68.3 & 62.5 & 58.0 & 65.5 & 90.0 & 85.0 & 84.0 & 73.0 & 83.0 \\
Gemini-3.1-Pro & 82.7 & 73.1 & 73.1 & 63.6 & 73.5 & 100.0 & 92.0 & 92.0 & 68.0 & 88.0 \\
DeepSeek-V4-Pro & 33.7 & 27.9 & 36.5 & 23.9 & 30.8 & 99.0 & 72.0 & 78.0 & 66.0 & 78.8 \\
Kimi-K2.6 & 46.2 & 28.8 & 36.5 & 35.2 & 36.8 & 94.0 & 48.0 & 58.0 & 34.0 & 58.5 \\
Qwen3.7-Max & 53.8 & 53.8 & 53.8 & 35.2 & 49.8 & 100.0 & 85.0 & 76.0 & 49.0 & 77.5 \\
Qwen3-Max & 32.7 & 21.2 & 36.5 & 27.3 & 29.5 & 100.0 & 65.0 & 65.0 & 49.0 & 69.8 \\
\midrule
MiniCPM4.1-8B & 8.7 & 2.9 & 4.8 & 5.7 & 5.5 & 75.0 & 24.0 & 12.0 & 15.0 & 31.5 \\
MiMo-Embodied-7B & 14.4 & 9.6 & 8.7 & 10.2 & 10.8 & 91.0 & 48.0 & 22.0 & 14.0 & 43.8 \\  
RynnBrain-8B & 11.5 & 5.8 & 5.8 & 6.8 & 7.5 & 70.0 & 36.0 & 11.0 & 13.0 & 32.5 \\
Qwen3-8B-NonThinking & 12.5 & 5.8 & 7.7 & 12.5 & 9.5 & 83.0 & 41.0 & 23.0 & 6.0 & 38.3 \\
Qwen3-8B-Thinking & 13.5 & 6.7 & 5.8 & 8.0 & 8.5 & 98.0 & 46.0 & 32.0 & 27.0 & 50.8 \\
Athena-SFT & 3.8 & 4.8 & 5.8 & 10.2 & 6.0 & 78.0 & 34.0 & 16.0 & 14.0 & 35.5 \\
\textbf{Athena-Brain-8B} & 74.0 & 51.9 & 57.7 & 52.3 & 59.3 & 64.0 & 44.0 & 38.0 & 29.0 & 43.8 \\
\bottomrule
\end{tabular}
\caption{Intermediate metric success rate (\%) by embodiment level and overall
score. State-Goal measures completion of the target future state, and State-Return
measures return to the requested past location.}
\label{tab:interactive_intermediate_by_level}
\end{table}

Table~\ref{tab:interactive_intermediate_by_level} reports the overall and embodiment-level results. Athena-Brain-8B substantially improves both intermediate probes over Athena-SFT, with a particularly large State-Goal gain from 6.0\% to 59.3\%. It also outperforms all other open-weight 7B--8B baselines on State-Goal by a wide margin, indicating that embodied expert training significantly strengthens the ability to execute short-horizon interaction objectives beyond what is reflected by end-to-end task success alone.

For State-Return, Athena-Brain-8B improves over the intermediate checkpoint and Qwen3-8B-NonThinking, and matches MiMo-Embodied-7B at 43.8\%. However, Qwen3-8B-Thinking reaches 50.8\%, compared with 38.3\% in non-thinking mode, while its State-Goal score decreases slightly from 9.5\% to 8.5\%. This contrast suggests that extended reasoning helps the base model recover past locations but does not translate into better near-future goal execution. Athena-Brain-8B therefore remains competitive among compact open-weight models on spatial recovery, although a noticeable gap remains relative to the strongest frontier models; long-range spatial memory and self-localization remain important bottlenecks.

\begin{table}[t]
\centering
\scriptsize
\setlength{\tabcolsep}{2.8pt}
\begin{tabular}{lccccccccc}
\toprule
\multirow{2}{*}{Model}
& \multicolumn{5}{c}{State-Goal (GSR)}
& \multicolumn{4}{c}{State-Return (RLSR)} \\
\cmidrule(lr){2-6} \cmidrule(lr){7-10}
& 0--5 & 6--10 & 11--20 & 21--40 & 41+
& 6--10 & 11--20 & 21--40 & 41+ \\
\midrule
GPT-5.5 & 54.8 & 62.7 & 65.9 & 68.4 & 66.9 & 89.7 & 87.2 & 80.0 & 71.8 \\
Gemini-3.1-Pro & 74.2 & 72.9 & 75.3 & 76.8 & 70.0 & 100.0 & 94.4 & 83.2 & 70.4 \\
DeepSeek-V4-Pro & 16.1 & 25.4 & 24.7 & 37.9 & 35.4 & 94.9 & 83.6 & 73.7 & 63.4 \\
Kimi-K2.6 & 12.9 & 39.0 & 37.6 & 37.9 & 40.0 & 82.1 & 69.7 & 43.2 & 35.2 \\
Qwen3.7-Max & 19.4 & 55.9 & 52.9 & 50.5 & 51.5 & 97.4 & 86.2 & 70.5 & 52.1 \\
Qwen3-Max & 0.0 & 27.1 & 32.9 & 29.5 & 35.4 & 97.4 & 77.9 & 58.9 & 46.5 \\
\midrule
MiniCPM4.1-8B & 0.0 & 8.5 & 8.2 & 8.4 & 1.5 & 64.1 & 40.5 & 12.6 & 14.1 \\
MiMo-Embodied-7B & 0.0 & 10.2 & 10.6 & 14.7 & 10.8 & 84.6 & 53.3 & 26.3 & 18.3 \\
RynnBrain-8B & 0.0 & 6.8 & 8.2 & 13.7 & 4.6 & 59.0 & 38.5 & 25.3 & 11.3 \\
Qwen3-8B-NonThinking & 0.0 & 10.2 & 8.2 & 15.8 & 7.7 & 79.5 & 50.8 & 15.8 & 11.3 \\
Qwen3-8B-Thinking & 0.0 & 10.2 & 14.1 & 7.4 & 6.9 & 92.3 & 63.1 & 31.6 & 19.7 \\
Athena-SFT & 0.0 & 8.5 & 4.7 & 6.3 & 6.9 & 76.9 & 42.1 & 18.9 & 16.9 \\
\textbf{Athena-Brain-8B} & 71.0 & 69.5 & 60.0 & 64.2 & 47.7 & 53.8 & 50.3 & 34.7 & 32.4 \\
\bottomrule
\end{tabular}
\caption{Intermediate metric success rate (\%) by query turn bin. State-Return
has no 0--5 bin under the current probe sampling.}
\label{tab:interactive_intermediate_turn_bins}
\end{table}

Table~\ref{tab:interactive_intermediate_turn_bins} further analyzes performance as a function of interaction progress. State-Goal is substantially less sensitive to interaction progress than State-Return. For both Qwen3-8B inference modes, State-Return is strongest at turns 6--10 and then declines as the history grows; thinking consistently improves recovery in every turn bin, but still falls from 92.3\% to 19.7\%. Athena-Brain-8B starts lower at 53.8\% but retains 32.4\% at turns 41+, the strongest late-episode State-Return result among the open-weight 7B--8B models. These trends indicate that recovering previously visited locations becomes increasingly difficult under accumulated partial observations.

Together, these intermediate probes complement end-to-end success rates by providing a more fine-grained view of embodied interaction capability. They show that Athena-Brain-8B's gains are driven most clearly by stronger goal-directed execution, while its comparatively robust late-episode State-Return performance points to improved long-horizon spatial recovery despite remaining headroom in overall return accuracy.

%% file: main/analysis.tex
\section{Analysis}

\subsection{Merge Analysis}

\begin{figure*}[t!]
\centering
\includegraphics[width=\textwidth]{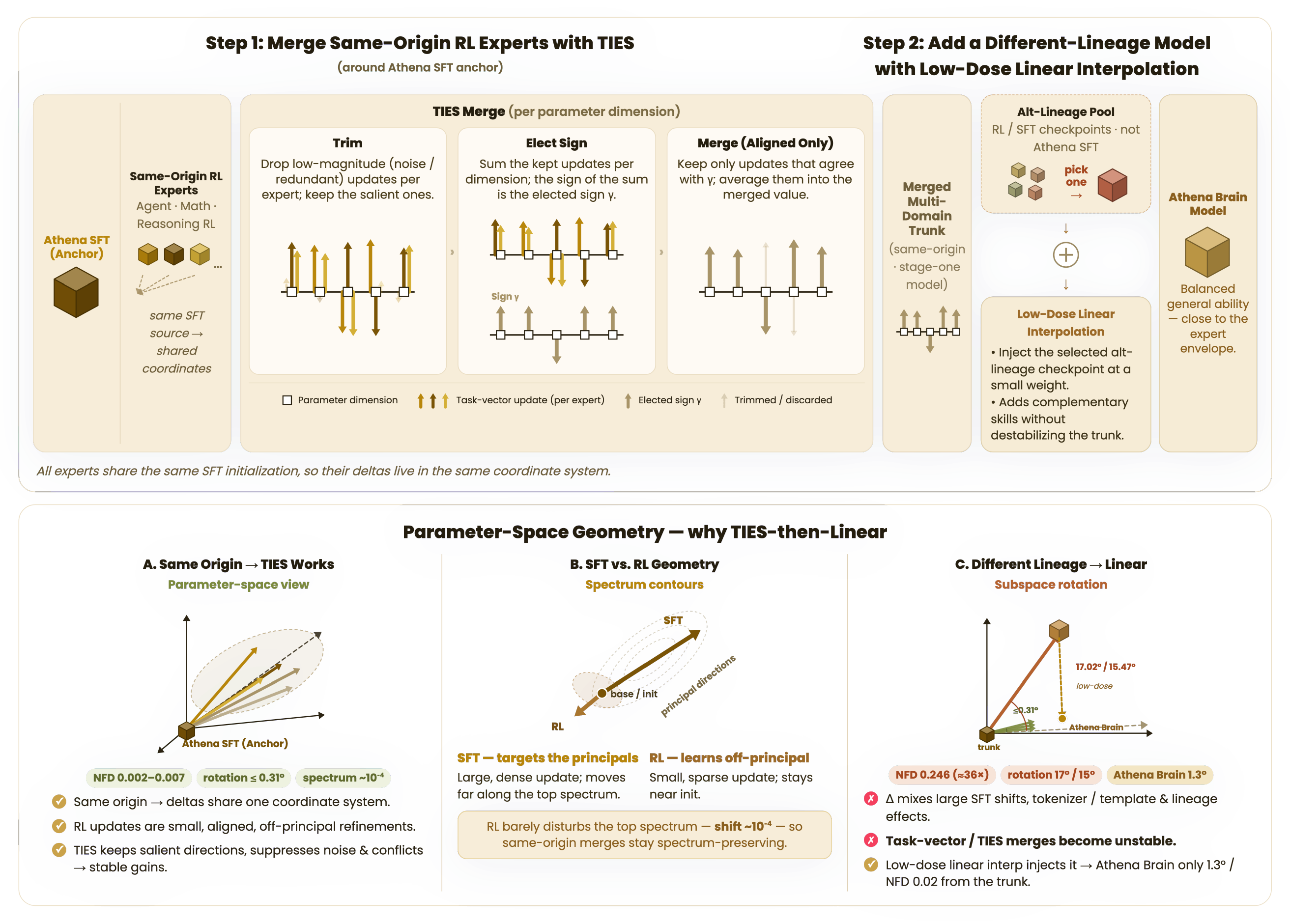}
\caption{Overview of the lineage-aware merge used to construct Athena-Brain-8B.
Same-origin RL experts are consolidated around the Athena SFT anchor with TIES,
while a complementary SFT prior from a different lineage is subsequently
introduced through a conservative linear correction. The lower panels
illustrate the parameter-space motivation for separating the two stages.}
\label{fig:athena_merge_overview}
\end{figure*}

We construct \textit{Athena-Brain-8B} with a lineage-aware, two-stage merge. The
first stage consolidates domain-specialized RL experts initialized from
\textit{Athena SFT}, while the second stage introduces a complementary post-trained checkpoint from a different training lineage through conservative linear
interpolation. Figure~\ref{fig:athena_merge_overview} summarizes the merge
procedure and the parameter-space motivation for separating these two update
regimes. This separation is deliberate: the two stages involve updates with
markedly different geometry and therefore call for different merge operators.

Prior work applies TIES to RL experts derived from a shared SFT
initialization~\citep{wang2026tomixmerge}. We extend this shared-lineage setting
to a heterogeneous-lineage merge. The first-stage experts define task vectors
relative to the same SFT anchor, making sign-based conflict resolution through
TIES well grounded. By contrast, the second-stage checkpoint comes from a
different lineage, so its displacement relative to the Athena SFT anchor
conflates lineage-level and task-specific differences. We therefore incorporate
it through a separate linear correction rather than treating it as another
expert in the same TIES vote. Our empirical results and update-geometry
analysis support this lineage-aware assignment of merge operators.

For a same-origin RL expert with parameters $\theta_i$, we define its task
vector relative to Athena SFT as
\[
\tau_i = \theta_i - \theta_{\mathrm{AthenaSFT}}.
\]
Because all experts share the same initialization, the vectors
$\{\tau_i\}_{i=1}^{N}$ are defined relative to a common anchor, so
coordinate-wise sign agreement can be interpreted as agreement among
post-training updates. We apply TIES~\citep{yadav2023ties} to
$\{\tau_i\}_{i=1}^{N}$, obtaining a merged update $\tau_{\mathrm{TIES}}$, and
construct the first-stage model as
\[
\theta_{\mathrm{stage1}}
=
\theta_{\mathrm{AthenaSFT}} + \tau_{\mathrm{TIES}}.
\]
TIES removes low-magnitude entries, elects an aggregate sign at each parameter
coordinate, and merges only entries aligned with that sign. In our setting,
this retains salient, sign-consistent updates while suppressing weak changes
and resolving cross-expert interference.

\begin{table}[th]
\centering
\small
\renewcommand{\arraystretch}{1.05}
\setlength{\tabcolsep}{3pt}
\begin{tabular}{@{\hspace{3pt}}lccc>{\hspace{0pt}}c<{\hspace{0pt}}@{\hspace{5pt}}}
\toprule
{Benchmark}
&
{Athena SFT}
&
{Best Expert}
&
{Athena-Brain-8B}
&
\shortstack{{$\Delta$(Athena-Brain-8B --}
                  {Expert)}} \\
\midrule

AIME 2025
& 64.58
& 70.00 {\tiny (RL)}
& \textbf{70.83}
& \textbf{+0.83} \\

AIME 2024
& 73.75
& 73.75 {\tiny (SFT)}
& \textbf{76.67}
& \textbf{+2.92} \\

LiveCodeBench v6
& 48.90
& 49.78 {\tiny (RL)}
& \textbf{54.63}
& \textbf{+4.85} \\

LiveBench
& 68.60
& \textbf{71.00} {\tiny (RL)}
& 70.80
& $-0.20$ \\

Zebra Logic
& 74.00
& 74.60 {\tiny (RL)}
& \textbf{76.70}
& \textbf{+2.10} \\

Arena-Hard v2
& 43.26
& 43.26 {\tiny (SFT)}
& \textbf{48.21}
& \textbf{+4.95} \\

GPQA Diamond
& 59.09
& 59.09 {\tiny (RL)}
& \textbf{59.60}
& \textbf{+0.51} \\

BFCL v1
& 82.66
& \textbf{83.09} {\tiny (RL)}
& 81.51
& $-1.58$ \\

BFCL v3
& 39.88
& 41.60 {\tiny (RL)}
& \textbf{43.13}
& \textbf{+1.53} \\

In-Domain Embodied Overall
& 6.98
& \textbf{60.11} {\tiny (embodied RL)}
& 58.52
& $-1.59$ \\

ALFWorld OOD\protect\footnotemark
& 16/134
& 47/134 {\tiny (embodied RL)}
& \textbf{49/134}
& \textbf{+2} \\

\bottomrule
\end{tabular}

\caption{Benchmark results of Athena-Brain-8B. The model types of the best
experts are tagged, including RL, SFT, and embodied RL.
$\Delta$ denotes the performance difference between Athena-Brain-8B and the
corresponding best expert.}
\label{tab:merge_best_expert}
\end{table}

\footnotetext{ALFWorld OOD is not used for targeted training in the merge
pipeline and is included as an out-of-distribution embodied-interaction probe.}

The same-origin RL experts remain in a
local neighborhood around Athena SFT, and the first-stage merged model preserves
this local structure. Their updates induce limited drift in the leading
singular spectrum and limited rotation of the corresponding singular
subspaces. This is consistent with prior observations that RLVR tends to update
off-principal, low-curvature, spectrum-preserving directions, whereas SFT
induces stronger principal-spectrum and subspace changes
\citep{zhu2025pathnot}.

The different-lineage SFT checkpoint exhibits a qualitatively different
geometry. Its displacement from both Athena SFT and the first-stage merged model
is substantially larger than the same-origin RL updates. When re-anchored at
the first-stage model, its update is less aligned with representative RL-derived directions and produces markedly stronger spectrum and subspace
changes. A shared ancestor alone is insufficient for TIES compatibility. Re-anchoring
the different-lineage SFT checkpoints at their common base model still produced
task vectors that were widely separated in parameter space and qualitatively
different in update geometry. TIES merging under this anchor caused severe
performance degradation, showing that formal anchor sharing does not guarantee
local comparability. Instead, coordinate-wise sign agreement becomes entangled
with large lineage-level shifts accumulated along different training paths.

We therefore incorporate the different-lineage checkpoint only after the TIES
stage, using low-weight linear interpolation. The resulting Athena-Brain-8B update
follows the intended correction direction while remaining much smaller than
the full source displacement. This preserves the local geometry established by
the same-origin RL merge while transferring complementary capability from the
second lineage. The two stages thus solve distinct composition problems: TIES
resolves interference among comparable RL task vectors, whereas conservative
linear interpolation introduces a distant prior without placing it in the same
voting pool.

Table~\ref{tab:merge_best_expert} compares Athena-Brain-8B with Athena SFT and
a relevant domain reference for each benchmark. Across the general benchmarks, Athena-Brain-8B delivers broad improvements over
Athena SFT while matching or surpassing the corresponding domain experts in most cases.

For embodied tasks, Athena-Brain-8B retains nearly all of the in-domain
performance of the embodied RL reference while substantially improving over
Athena SFT. On ALFWorld, which is not used for targeted training in this pipeline, Athena-Brain-8B exceeds the embodied RL reference. These results
show that the two-stage construction combines most reference-expert
capabilities in a single checkpoint, while the preceding geometry analysis
explains why the two update regimes are handled separately.

\subsection{Rollout Budget Analysis}
\label{sec:rollout_bugdet}

\begin{figure}[t]
\centering
\begin{subfigure}{0.32\linewidth}
\centering
\includegraphics[width=\linewidth]{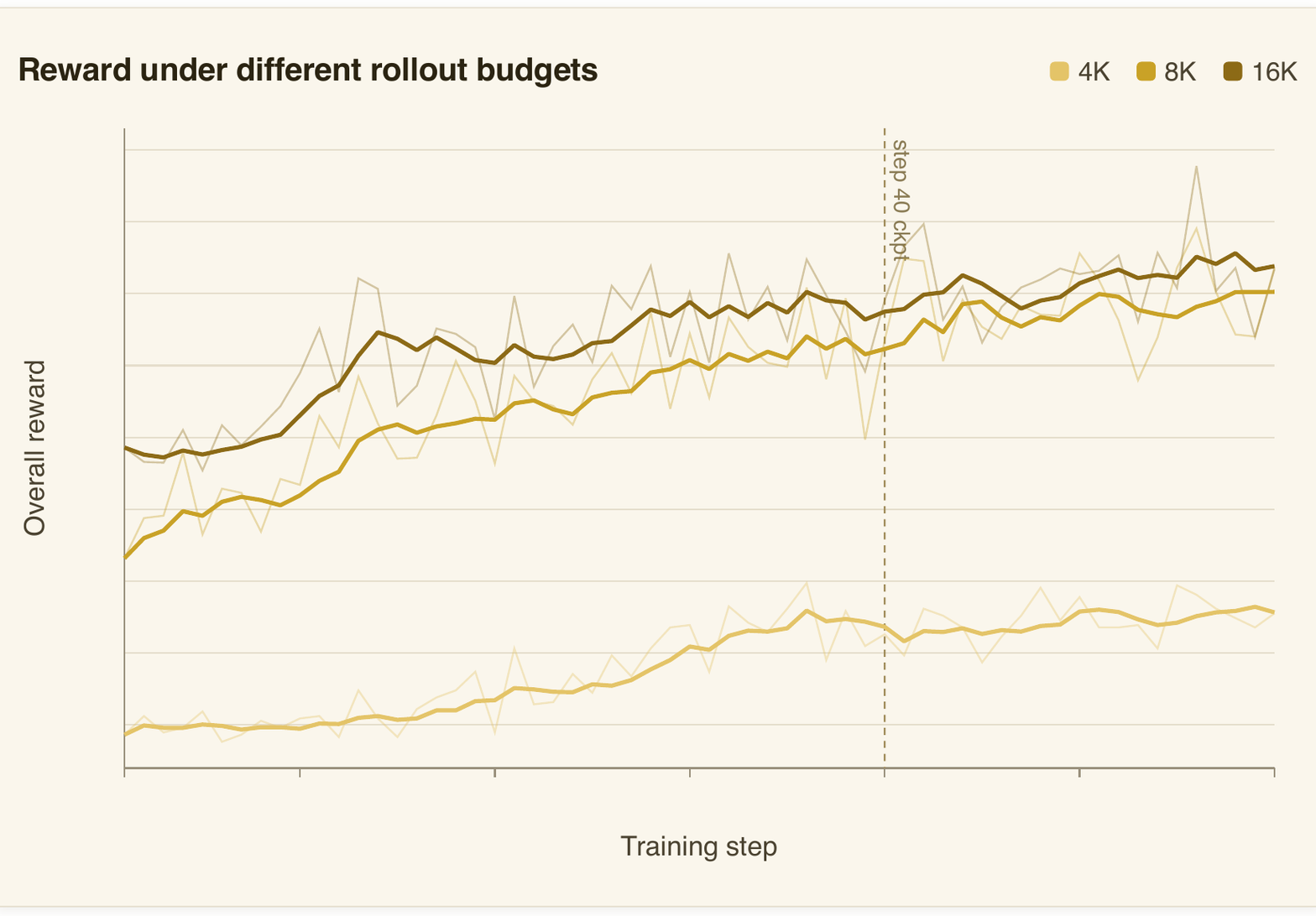}
\caption{Training reward under different rollout budgets.}
\label{fig:3exp_reward}
\end{subfigure}
\hfill
\begin{subfigure}{0.32\linewidth}
\centering
\includegraphics[width=\linewidth]{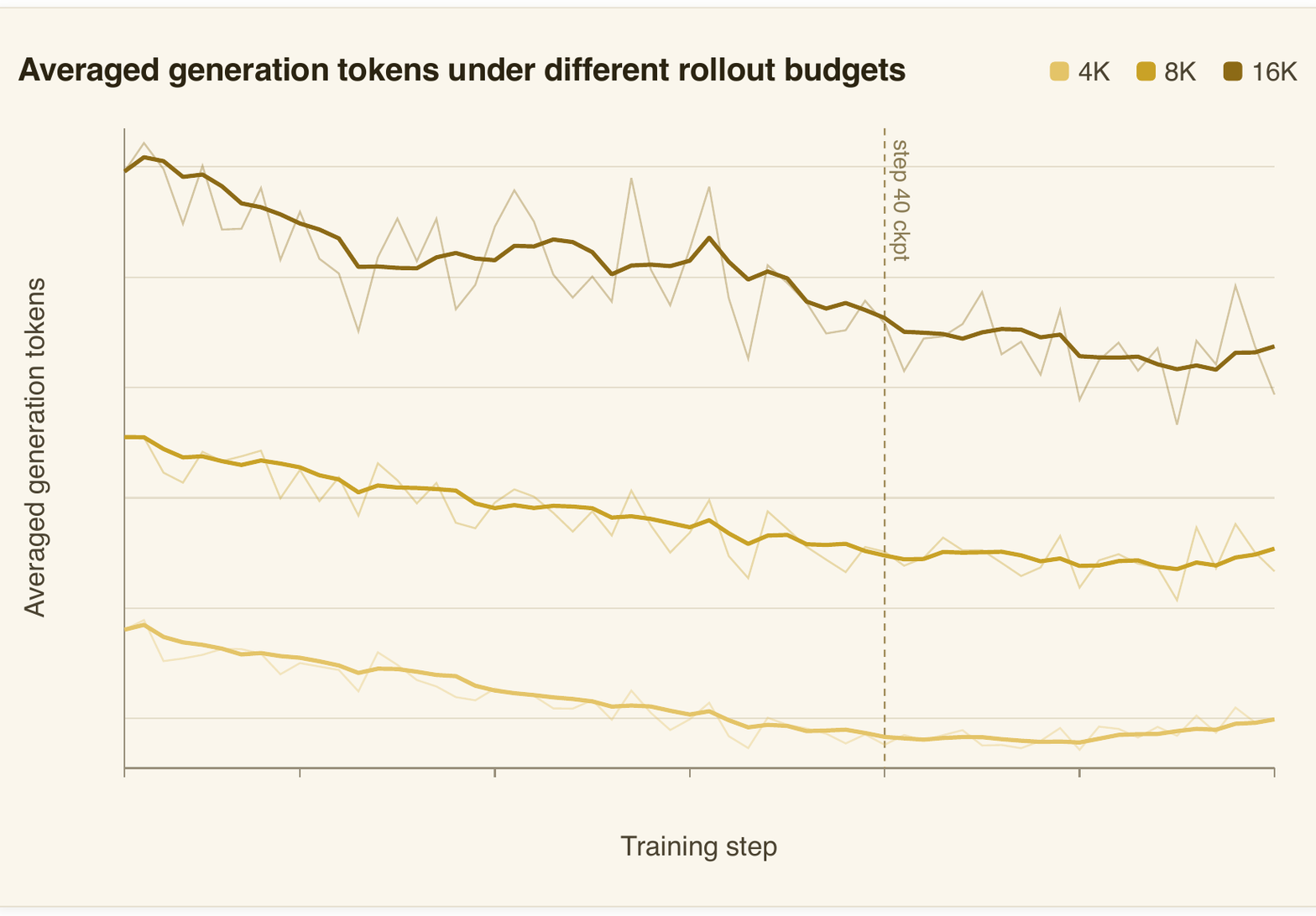}
\caption{Average generation length during training.}
\label{fig:3exp_generation}
\end{subfigure}
\hfill
\begin{subfigure}{0.32\linewidth}
\centering
\includegraphics[width=\linewidth]{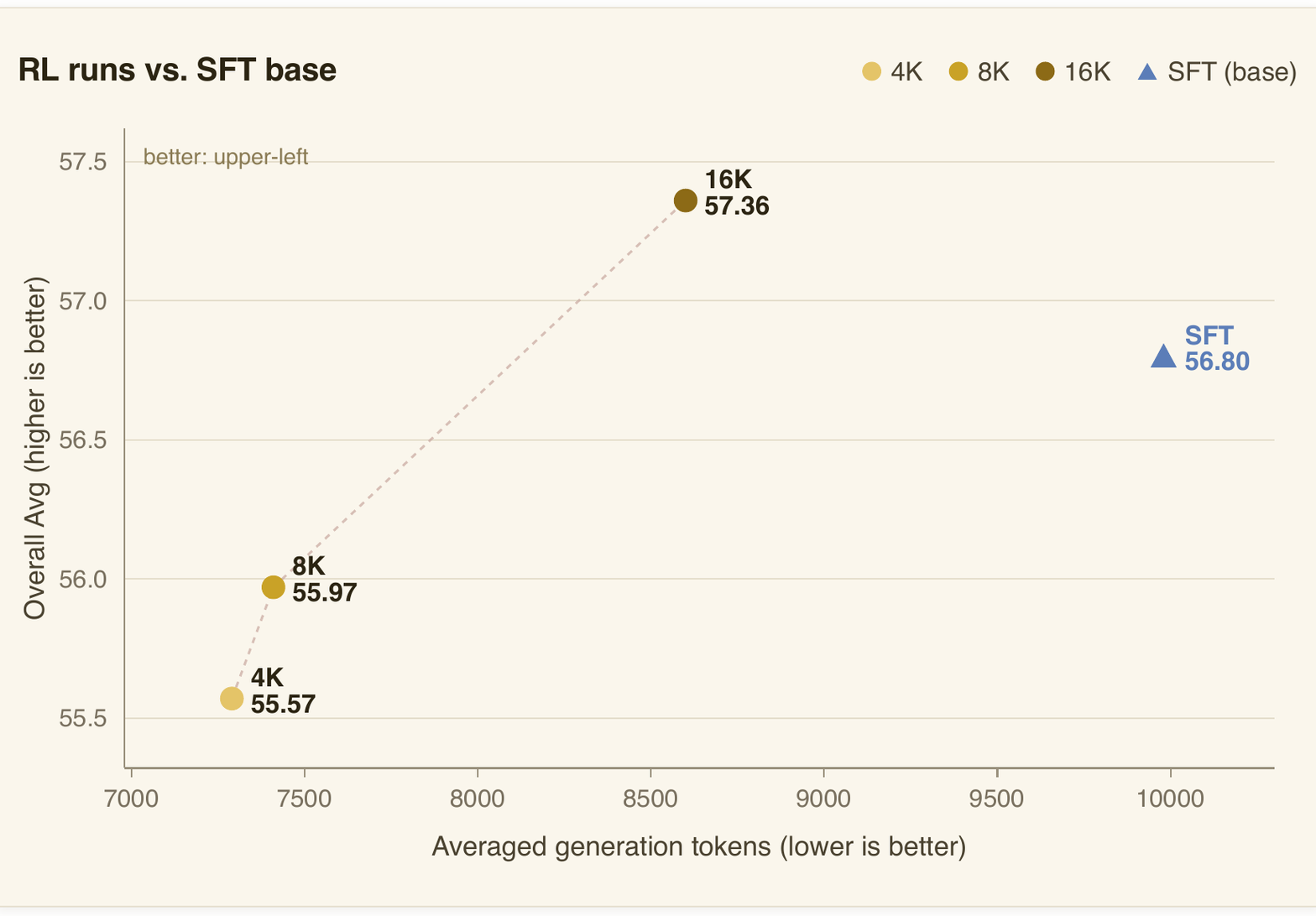}
\caption{Downstream performance vs. generation length (step-40 checkpoints).}
\label{fig:3_exp_res}
\end{subfigure}
\caption{
Effect of the maximum rollout budget (4K, 8K, 16K tokens) during General RL. (a) Training reward curves: all three settings converge after approximately 40 optimization steps. (b) Average generation length over training: larger budgets sustain longer responses throughout optimization. (c) Downstream reasoning performance versus average generation length, evaluated with the step-40 checkpoint of each setting; the SFT baseline is shown for reference. Points toward the upper-left indicate a better capability–efficiency trade-off.
}
\label{fig:rollout_budget}
\end{figure}

The maximum rollout budget during General RL constrains how long sampled responses can grow, affecting both the reasoning behaviors that can be rewarded and the strength of the token-budget reward. We train Athena-SFT on the same small mixed RL dataset under three budgets (4K, 8K, and 16K tokens). As shown in Figure~\ref{fig:rollout_budget} (a), all settings are about to converge stably after approximately 40 optimization steps, and we use the step-40 checkpoints for comparison. Figure~\ref{fig:rollout_budget} (b) shows that generation length decreases under all settings driven by the token-budget reward, while the achievable length remains capped by the budget.

Figure~\ref{fig:rollout_budget} (c) compares downstream performance against average generation length. Restrictive budgets clearly hurt: the 4K setting substantially degrades performance on complex reasoning tasks, whereas increasing the budget to 16K consistently improves performance with only a moderate growth in generation length. We do not extend the budget further, since an excessively large budget would render the token-budget reward ineffective as a length regularizer (Section 3.2). We therefore adopt a 16K rollout budget for General RL: large enough to preserve reasoning quality on hard tasks, yet tight enough to keep generation concise.

\subsection{Ablation Study of Embodied RL}
\label{sec:embodied-rl-ablation}

We further ablate the main design choices in embodied RL using balanced Museum--Supermarket training data. The ablations vary three axes: truncation, reward design, and map sampling.

\paragraph{Truncation Rule.}
The \textit{all-error-stop} rule terminates a rollout on any tool or execution error, including both protocol-level formatting failures and executable tool failures. In contrast, the \textit{format-only-stop} rule terminates only on format-level protocol errors and treats unknown tools, locked tools, and executable skill failures as recoverable, allowing the rollout to continue after valid but unsuccessful actions.

\paragraph{Reward Scheme.}
The \textit{punitive} scheme assigns a terminal reward of $-100$ for truncation and a $-10$ tool-error penalty, making tool failures a strong negative signal during optimization. The \textit{mild} scheme removes the extra truncation penalty and reduces the tool-error penalty to $-0.1$, preserving a lightweight constraint on invalid actions while keeping the reward signal focused on task progress.

\paragraph{Map Sampling.}
Full-map training uses all training maps, exposing the model to the complete range of spatial layouts and task complexity. Partial-map training uses the easier subset for all embodiment levels, reducing the effective search difficulty while preserving the same environment types.

\begin{table}[t]
\centering
\scriptsize
\setlength{\tabcolsep}{4pt}
\begin{tabular}{llllccc}
\toprule
Model & Map data & Reward scheme & Truncation rule & Museum & Supermarket & Overall \\
\midrule
Pre-RL baseline & -- & -- & -- & 0.5125 & 0.5162 & 0.5145 \\
\midrule
\multirow{8}{*}{Embodied RL} 
& Full & Punitive & Format-only-stop & 0.5675 & 0.5375 & 0.5511 \\ 
& Full & Punitive & All-error-stop & 0.5975 & 0.5000 & 0.5443 \\ 
& Full & Mild & All-error-stop & 0.5675 & 0.5292 & 0.5466 \\ 
& Full & Mild & Format-only-stop & 0.5800 & 0.5333 & 0.5545 \\ 
\cmidrule(l){2-7}
& Partial & Punitive & Format-only-stop & 0.5575 & 0.5292 & 0.5421 \\
& Partial & Punitive & All-error-stop & 0.5850 & 0.5208 & 0.5500 \\
& Partial & Mild & All-error-stop & 0.5600 & 0.5625 & 0.5614 \\
& Partial & Mild & Format-only-stop & \textbf{0.6125} & \textbf{0.5896} & \textbf{0.6000} \\
\bottomrule
\end{tabular}
\caption{Embodied RL ablation results.}
\label{tab:embodied_rl_ablation}
\end{table}

Table~\ref{tab:embodied_rl_ablation} shows that the best-balanced performance is achieved with partial maps, the mild reward scheme, and format-only-stop truncation. This configuration attains a weighted overall score of 0.6000, improving over the pre-RL baseline by 0.1000 on Museum, 0.0734 on Supermarket, and 0.0855 overall. It is also the only variant that yields substantial gains in both environments simultaneously, indicating that aligning penalty scale and truncation with closed-loop interaction is critical for embodied RL.

Reward design has the clearest effect on Supermarket. Under partial maps and the all-error-stop truncation rule, switching from the punitive reward scheme to the mild scheme increases the best Supermarket score from 0.5208 to 0.5625 and the weighted overall score from 0.5500 to 0.5614. This suggests that overly large tool and truncation penalties can make RL over-optimize protocol avoidance, while a milder penalty better preserves task-progress learning and recovery from executable mistakes.

The truncation policy and map sampling further interact with this reward choice. With the mild reward scheme, the format-only-stop rule gives the strongest overall result, indicating that recoverable tool and execution errors can provide useful recovery trajectories instead of simply ending the rollout. Full-map training also improves over the pre-RL baseline, with its best variant reaching 0.5545 overall, but it remains clearly below the best partial-map result of 0.6000. The gap mainly comes from Supermarket, where full-map training introduces higher exploration variance and reaches only 0.5333 in its best format-only-stop setting, compared with 0.5896 under partial maps. This indicates that the training curriculum should maintain a moderate difficulty level: overly difficult or high-variance maps can weaken the learning signal and lead to worse policy improvement, even when they contain more environmental diversity.

Beyond the final evaluation metrics reported above, we further analyze the training dynamics to understand how different configurations affect the learning process over time.
Since the punitive and mild reward schemes operate on different reward scales, we do not directly compare reward values but instead use task completion rate and tool protocol error rate as the primary metrics. Statistics are reported at Step~100, Step~200, and Step~300.

\paragraph{Partial Maps.}

\begin{table}[t]
\centering
\scriptsize
\setlength{\tabcolsep}{4pt}
\begin{tabular}{llcccccc}
\toprule
Truncation & Reward & Comp.@100 & Comp.@200 & Comp.@300 & Tool Err.@100 & Tool Err.@200 & Tool Err.@300 \\
\midrule
Format-only-stop & Punitive & 0.562 & 0.612 & 0.704 & 0.014 & 0.015 & 0.014 \\
All-error-stop & Punitive & 0.500 & 0.530 & 0.800 & 0.321 & 0.076 & 0.091 \\
All-error-stop & Mild & 0.529 & 0.630 & 0.865 & 0.343 & 0.074 & 0.019 \\
Format-only-stop & Mild & 0.623 & 0.605 & 0.847 & 0.000 & 0.000 & 0.000 \\
\bottomrule
\end{tabular}
\caption{Training dynamics on partial maps. Comp.\ denotes task completion rate; Tool Err.\ denotes tool protocol error rate.}
\label{tab:partial_map_dynamics}
\end{table}

On partial maps, the mild reward consistently outperforms the punitive reward under both truncation schemes: Step~300 completion improves from $0.704$ to $0.847$ (format-only-stop) and from $0.800$ to $0.865$ (all-error-stop), without increasing tool errors. The mild reward also accelerates early-stage learning. Although all-error-stop achieves slightly higher Comp@300 than format-only-stop (0.865 vs. 0.847), the final evaluation tells a different story: format-only-stop with mild reward reaches 0.6000 overall, substantially outperforming all-error-stop (0.5614). This suggests that higher training-time completion does not necessarily translate into a better final policy; preserving recovery trajectories under format-only-stop yields stronger generalization even in simpler environments.

\emph{Partial map conclusion:} Mild reward is the primary improvement driver. While all-error-stop shows slightly higher training-time completion, format-only-stop produces a substantially better final policy, indicating that preserving recoverable errors benefits policy quality regardless of environment complexity.
\paragraph{Full Maps.}

\begin{table}[t]
\centering
\scriptsize
\setlength{\tabcolsep}{4pt}
\begin{tabular}{llcccccc}
\toprule
Truncation & Reward & Comp.@100 & Comp.@200 & Comp.@300 & Tool Err.@100 & Tool Err.@200 & Tool Err.@300 \\
\midrule
Format-only-stop & Punitive & 0.580 & 0.554 & 0.814 & 0.000 & 0.012 & 0.014 \\
All-error-stop & Punitive & 0.303 & 0.463 & 0.566 & 0.289 & 0.150 & 0.000 \\
All-error-stop & Mild & 0.234 & 0.541 & 0.550 & 0.516 & 0.216 & 0.125 \\
Format-only-stop & Mild & 0.563 & 0.635 & 0.718 & 0.014 & 0.000 & 0.000 \\
\bottomrule
\end{tabular}
\caption{Training dynamics on full maps.}
\label{tab:full_map_dynamics}
\end{table}

On full maps, the truncation effect reverses: format-only-stop substantially outperforms all-error-stop, with Step~300 completion gaps of $0.248$ (punitive) and $0.168$ (mild)---far larger than on partial maps. All-error-stop suffers severely in early training (Step~100 completion as low as $0.234$), and although tool errors decrease over time, completion rates remain far below format-only-stop. The reward scheme shows inconsistent effects: punitive slightly outperforms mild under both truncation rules, indicating that reducing truncation penalty alone is insufficient---the main bottleneck is spatial planning and long-horizon decision making. Format-only-stop preserves recovery trajectories after decision errors (path selection, tool choice, state judgment), providing more effective training signals in complex environments.

\emph{Full map conclusion:} Format-only-stop is the primary improvement driver; preserving error recovery trajectories matters more than strengthening truncation constraints.

\paragraph{Summary.}

Partial and full maps exhibit different patterns in degree but converge on the same conclusion. On both map types, format-only-stop produces the best final policy, confirming that preserving error recovery trajectories is universally beneficial. The difference lies in magnitude: on partial maps, the gap between truncation strategies is smaller in training-time completion but still decisive in final evaluation; on full maps, the gap is already large during training. Mild reward consistently helps on partial maps, while its effect on full maps is weaker, suggesting that reward scale matters more in simpler environments. Overall, these results support format-only-stop with mild reward as the preferred configuration across complexity levels.

\subsection{Case Analysis}

\begin{figure}[tbp]
    \centering \includegraphics[width=1\linewidth]{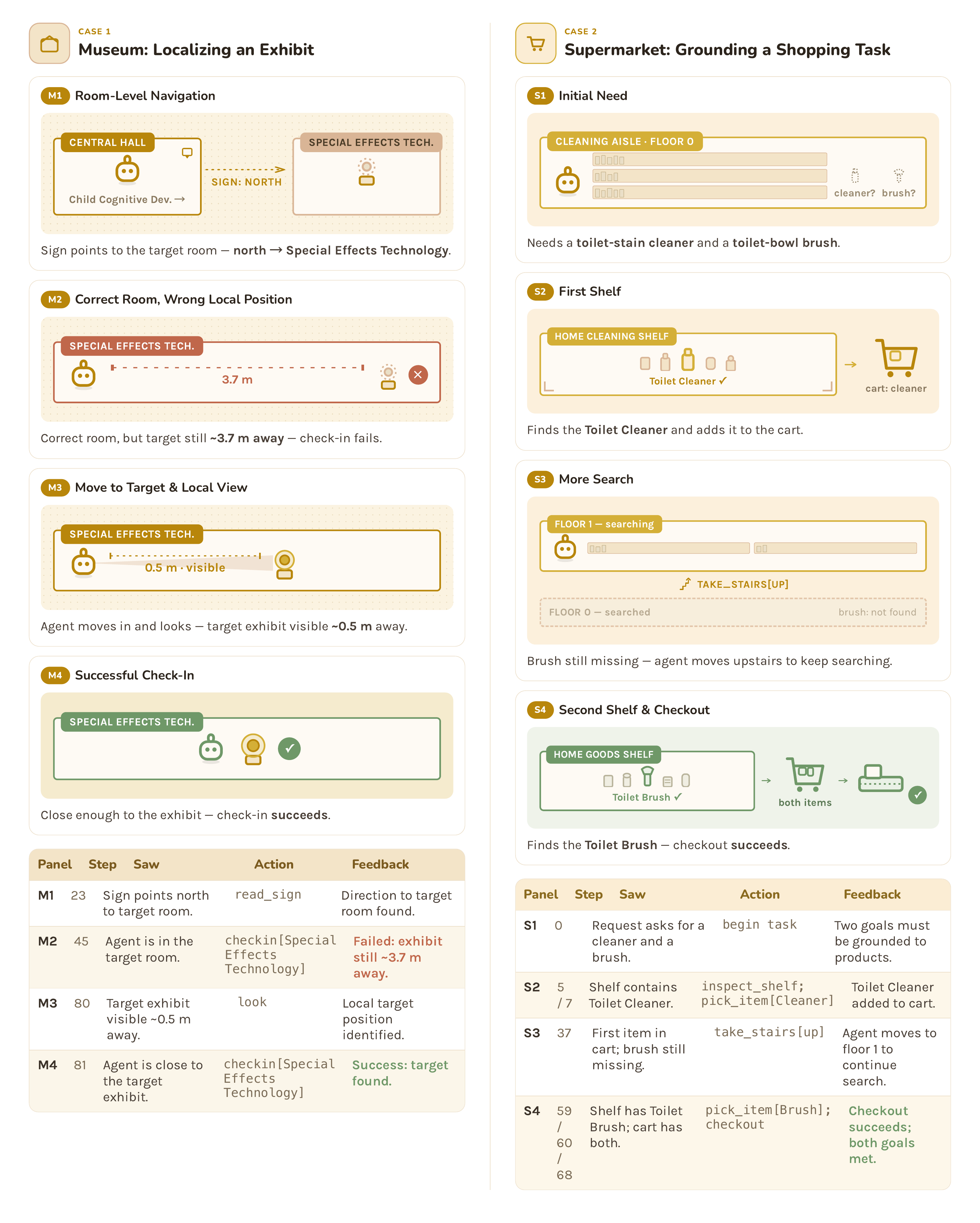}
    \caption{Qualitative examples of embodied task solving. Left: the agent uses local visual feedback to correct its position after reaching the target museum room but failing check-in. Right: the agent grounds a functional shopping request into concrete products, searches across shelves and floors, and completes checkout.}
    \label{fig:cases}
\end{figure}

\paragraph{Case Analysis.}
Figure~\ref{fig:cases} presents two representative interaction trajectories from the museum and supermarket environments.

The museum example illustrates the model's ability to recover from execution errors during long-horizon interaction. After an unsuccessful \texttt{checkin} attempt, the model does not repeatedly invoke the same action, but instead infers that its current position is incorrect, performs additional navigation to refine its location, and successfully completes the task. This behavior demonstrates that the policy can utilize environmental feedback to revise its spatial hypothesis and recover from intermediate failures.

The supermarket example highlights a more challenging form of task-oriented planning. Instead of specifying product names, the user request describes functional requirements: purchasing a cleaner for dissolving toilet stains together with a brush for cleaning the inside of a toilet bowl. The model must first infer the appropriate products from these descriptions, recognize that the two items are functionally complementary, and then plan an efficient shopping strategy. In this Level-2 episode, Athena-Brain-8B successfully completes a 68-step cross-floor shopping task, inspecting eight shelves before collecting both target items and finishing checkout. The trajectory demonstrates that the model integrates semantic understanding, long-horizon planning, and spatial exploration, rather than merely following explicit product names or predefined action sequences.

%% file: main/related_work.tex
\section{Related Work}
\label{sec:related}

\paragraph{Compact Language Models for Interactive Intelligence.}
Recent advances in post-training have substantially improved the reasoning capability of compact language models through supervised fine-tuning and reinforcement learning, enabling 8B-scale models to achieve increasingly competitive performance while remaining suitable for efficient deployment \citep{ouyang2022training,bai2022constitutional,wei2022chain,cobbe2021training,lightman2023lets,touvron2023llama,dubey2024llama,yang2025qwen3technicalreport}. In parallel, coding agents have demonstrated that language models can solve complex software tasks through long-horizon interaction involving iterative tool use, execution monitoring, and self-correction \citep{yang2024swe,wang2025openhands}. Although software and physical interaction share many common characteristics, existing coding agents primarily target cloud-based software environments and are not designed for the efficiency and latency requirements of on-device robotic systems. Athena-Brain-8B builds upon compact reasoning models while extending interactive intelligence toward resource-constrained embodied deployment.

\paragraph{Robot Brains and Embodied Intelligence.}
Large language models and vision-language models have increasingly been adopted as high-level controllers for robotic systems, enabling language-guided planning, skill composition, and interaction with physical environments \citep{brohan2023can,driess2023palm,zitkovich2023rt,kim2025openvla}. These systems tightly integrate foundation models with perception, navigation, manipulation, and low-level control, significantly advancing embodied intelligence. In contrast, Athena-Brain-8B focuses on the robot brain itself---a compact cognitive model responsible for reasoning, memory, planning, and tool use. By operating over structured observations and executable tools instead of raw sensory inputs or motor commands, Athena-Brain-8B remains modular while serving as a general cognitive layer that can be integrated with diverse perception and control systems.

\paragraph{Interactive Training Environments.}
Interactive environments provide scalable platforms for studying long-horizon decision making under partial observability. Text-based environments such as TextWorld and ALFWorld require agents to infer hidden state from textual observations, execute valid actions, and complete tasks through multi-step interaction \citep{barnestextworld,shridhar2021alfworld}. Inspired by this paradigm, we use text-based embodied environments as a scalable proxy for robot-brain post-training rather than a replacement for physical simulation or real-robot evaluation. Compared with prior environments, our platform introduces a unified tool-mediated interaction interface together with a progressive embodiment curriculum that systematically increases spatial reasoning complexity while maintaining consistent task semantics. This design enables scalable post-training of high-level embodied cognition and reproducible evaluation through executable interaction and programmatic verification.

%% file: main/conclusion.tex
\section{Conclusion and Future Work}
\label{sec:conclusion}

In this technical report, we presented \textbf{Athena-Brain-8B}, a compact 8B robot-brain model designed for efficient on-device deployment. Through a unified post-training pipeline combining supervised fine-tuning, reinforcement learning, embodied expert training, and model merging, Athena-Brain-8B integrates strong general language capability, efficient reasoning, and robust embodied interaction into a single compact checkpoint. Extensive experiments demonstrate that the model achieves competitive general capability with 7B/8B models while substantially improving embodied interaction performance, suggesting that compact language models can serve as practical robot brains.

Despite these encouraging results, Athena-Brain-8B represents only an initial step toward deployable embodied intelligence. Our current embodied environments capture key properties of robot interaction, including partial observability, tool-mediated actions, and long-horizon decision making, but remain limited in scale and diversity. Future work will expand these environments to more tasks, richer object interactions, and different robot embodiments, while also bridging the gap to real-world systems through simulation-based training, visual-language perception, and deployment on physical robots.

Beyond scaling environments, we believe future robot brains should move beyond conventional step-based interaction. Real-world environments evolve continuously, requiring agents to reason over asynchronous observations, persistent actions, and dynamic events. We therefore plan to explore the \emph{Engagement Process} framework~\citep{li2026engagementprocessrethinkingtemporal}, which models interaction as continuous engagement with the environment rather than discrete observation--action loops.

Finally, we envision Athena-Brain-8B evolving toward a unified interactive intelligence model spanning both physical and software environments. Recent advances in coding agents suggest that long-horizon planning, iterative tool use, execution monitoring, and error recovery learned in software environments can naturally complement embodied interaction. Conversely, executable code can itself become a powerful action interface for robot brains, enabling language models to dynamically synthesize behaviors beyond predefined tool APIs. We view Athena-Brain-8B as a compact foundation for this broader vision of integrating reasoning, coding, and embodied interaction into a unified robot brain.